\documentclass[lettersize,journal]{IEEEtran}
\usepackage{amsmath,amsfonts}
\usepackage{algorithmic}
\usepackage{array}
\usepackage[caption=false,font=normalsize,labelfont=sf,textfont=sf]{subfig}
\usepackage{textcomp}
\usepackage{stfloats}
\usepackage{url}
\usepackage{verbatim}
\usepackage{graphicx}

\usepackage{cite}
\usepackage{amsmath,amssymb,amsfonts}
\usepackage{algorithmic}
\usepackage{graphicx}
\usepackage{textcomp}
\usepackage{xcolor}

\usepackage{booktabs}
\usepackage{enumitem}
\usepackage{float}
\usepackage{mathrsfs}
\usepackage{bbding}
\usepackage{caption}
\usepackage{subcaption}
\usepackage{url}
\usepackage{xcolor}
\usepackage{calligra}
\usepackage[linesnumbered,ruled,lined,boxed]{algorithm2e}
\usepackage{multirow}
\usepackage[misc]{ifsym}
\newcommand{\bluecallig}[1]{{\scalebox{1.02}{\textbf{\textit{#1}}}}}

\newcommand{\syx}[1]{{\color{black} #1}}

\def\BibTeX{{\rm B\kern-.05em{\sc i\kern-.025em b}\kern-.08em
    T\kern-.1667em\lower.7ex\hbox{E}\kern-.125emX}}
\usepackage{balance}
\begin{document}

\title{Scene-Driven Multimodal Knowledge Graph Construction for Embodied AI
}

\author{
\IEEEauthorblockN{
Yaoxian Song\textsuperscript{1}\IEEEauthorrefmark{1}, \thanks{\IEEEauthorrefmark{1} Equal contribution}
Penglei Sun\textsuperscript{1}\IEEEauthorrefmark{1},
Haoyu Liu\textsuperscript{2},
Zhixu Li\textsuperscript{1}\textsuperscript{\Letter}, 
Wei Song\textsuperscript{2}\textsuperscript{\Letter},\\
Yanghua Xiao\textsuperscript{1},
and Xiaofang Zhou\textsuperscript{3}, Fellow, IEEE
}


\IEEEauthorblockA{\textsuperscript{1} Shanghai Key Laboratory of Data Science, School of Computer Science, Fudan University, China\\
aaronskw236@gmail.com,\{plsun20,zhixuli,shawyh\}@fudan.edu.cn}\\
\IEEEauthorblockA{\textsuperscript{2} Research Center for Intelligent Robotics, Zhejiang Lab, China,\\
hyuliu20@gmail.com, weisong@zhejianglab.com}\\
\IEEEauthorblockA{\textsuperscript{3} The Hong Kong University of Science and Technology, Hongkong SAR, China,\\
zxf@cse.ust.hk}
}

\markboth{Journal of \LaTeX\ Class Files,~Vol.~18, No.~9, September~2020}%
{How to Use the IEEEtran \LaTeX \ Templates}

\maketitle

\begin{abstract}
Embodied AI is one of the most popular studies in artificial intelligence and robotics, which can effectively improve the intelligence of real-world agents (i.e. robots) serving human beings.
Scene knowledge is important for an agent to understand the surroundings and make correct decisions in the varied open world. Currently, knowledge base for embodied tasks is missing and most existing work use general knowledge base or pre-trained models to enhance the intelligence of an agent. For conventional knowledge base, it is sparse, insufficient in capacity and cost in data collection. For pre-trained models, they face the uncertainty of knowledge and hard maintenance. To overcome the challenges of scene knowledge, we propose a scene-driven multimodal knowledge graph (Scene-MMKG) construction method combining conventional knowledge engineering and large language models. A unified scene knowledge injection framework is introduced for knowledge representation. To evaluate the advantages of our proposed method, we instantiate Scene-MMKG considering typical indoor robotic functionalities (\textbf{Manip}ulation and\textbf{ Mob}ility), named \textbf{ManipMob-MMKG}. Comparisons in characteristics indicate our instantiated ManipMob-MMKG has broad superiority on data-collection efficiency and knowledge quality. Experimental results on typical embodied tasks show that knowledge-enhanced methods using our instantiated ManipMob-MMKG can improve the performance obviously without re-designing model structures complexly. Our project can be found on \url{https://sites.google.com/view/manipmob-mmkg}
\end{abstract}

\begin{IEEEkeywords}
Multimodal Knowledge Graph, Scene Driven, Embodied AI, Robotic Intelligence
\end{IEEEkeywords}

\section{Introduction}\label{sec:introduction}

\begin{figure*}[htp]
\centering
\vspace{-0.3cm}  
\includegraphics[width=0.9\linewidth]{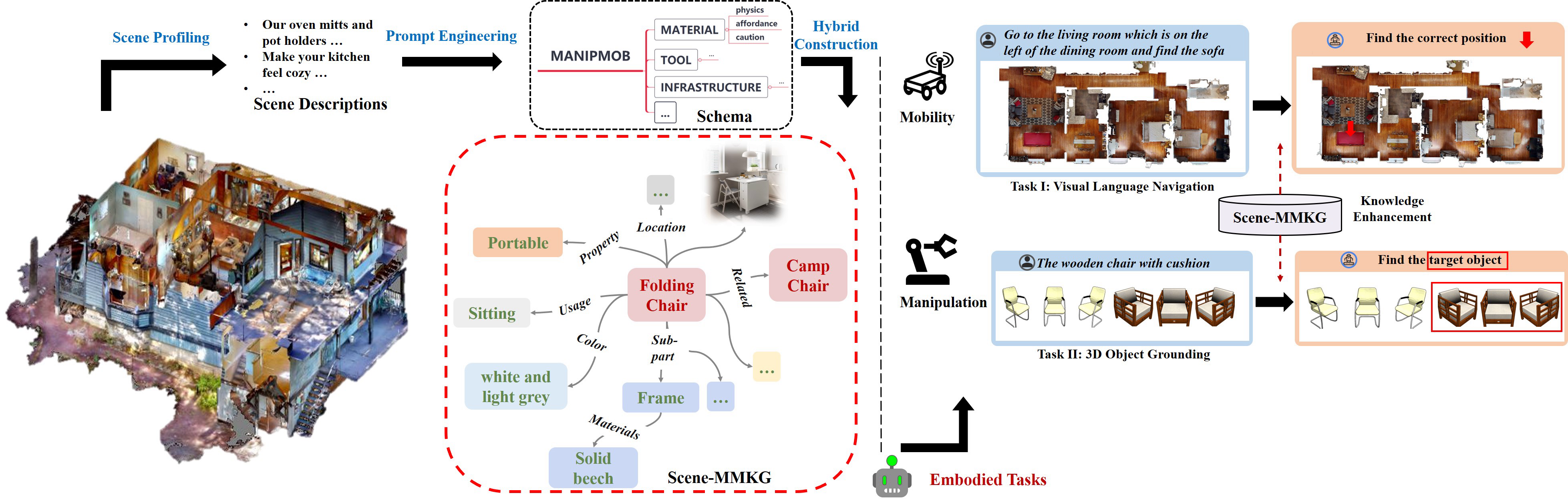}
\caption{The illustration of Scene-MMKG construction and embodied tasks using Scene-MMKG. Given the instruction and multimodal perception, \syx{the agent is required to retrieve knowledge} and answer the ``where" and ``which" question.}
\label{fig:overall}
\end{figure*}

Embodied Artificial Intelligence (AI) studies the intelligent behavior of an agent through its interaction with the environment~\cite{henderson2003human,gupta2021deep,patterson2007you}.
An agent can be some form of robot or any intelligent device that, like humans, learns and completes assigned tasks by observing and interacting with the environment. 
The environment, either in a physical or virtual scenario, contains stuff, human, and other agents in the scenario.
Thus, the knowledge about the environment, i.e., scene knowledge, greatly determines the level of intelligence in embodied AI.



%
%


\syx{Inspired by the human consciousness research~\cite{myllyla2022expertise} and the characteristics of embodied AI}, the scene knowledge required by an embodied agent could be roughly divided into two groups: 1) {\em Perceptual Knowledge} which describes the perceptible properties (such as object type, pose, location) of stuff, human and agent, as well as the relationships (such as relative position) between them in the scenario.
2) {\em Apperceptive Knowledge} that includes concept, fact, commonsense, affordance, emotion, intention, etc. about the human, agent, and stuff in the scenario.
%
%
While perceptual information could be gotten from the scenario with conventional pattern-recognition-based technologies, apperceptive knowledge can only be obtained from external resources.
For instance, a pre-trained robotic grasping detection model containing perceptual visual knowledge could generate a stable pose to let a robot grasp a target object (i.e. \textit{Mug}). But if apperceptive knowledge like affordance (i.e. [\textit{Mug}, hasUse, \textit{drinking water}]) is given, a robot needs to grasp \textit{Mug} horizontally by handle instead rim of the \textit{Mug}, avoiding water spilling from the mug or fingertip polluting the rim during grasping detection.

Conventional embodied AI efforts obtain scene knowledge in two ways.
The first way utilizes symbolic knowledge from various existing knowledge bases (KBs) such as uni-modal knowledge bases like WordNet~\cite{miller1995wordnet}, ConceptNet~\cite{liu2004conceptnet} and Wikidata~\cite{vrandevcic2014wikidata} for apperceptive knowledge, or multimodal knowledge bases~\cite{zhu2022multi,deng2023construction} like ImageNet~\cite{deng2009imagenet}, Visual Genome (VG)~\cite{krishna2017visual}, etc. for perceptual knowledge.
However, for scene knowledge used in embodied tasks\syx{~\cite{liu2023survey}}, the existing knowledge bases are sparse in details, imbalanced in categories, and small in capacity. It is also impractical to solve the flaws to construct a large and complete dataset with a limited cost.
The second way is to store knowledge using learnable parameters in pre-trained models (PTM). 
For instance, pre-trained models (i.e. YOLO~\cite{xu2022pp}) for object detection and scene segmentation can provide various scene knowledge about visual semantics. 
Pre-trained for image caption, video caption, and scene graph generation~\cite{xu2015show,tang2002spatial,johnson2015image} can generate cross-modal apperceptive knowledge in text or graph passively.
An agent can obtain scene knowledge using pre-trained visual question answering and visual language grounding~\cite{ma2022sqa3d,yu2016modeling} by human-robot interaction actively. However, most conventional pre-trained models contain issues in robustness and domain generalization for scene-changing embodied tasks.


For that, recently, with the booming of large pre-trained models (also named foundation model), it shows impressive performance on scene understanding and reasoning in zero-shot benefiting from internet-scale training data. ChatGPT~\cite{vemprala2023chatgpt,ChatGPTO40:online} and GPT-4~\cite{openai2023gpt4} contain a mass of parameter knowledge and have achieved excellent performance in large numbers of benchmark tasks, which indicates plausibly they are very close to passing the Turing test. For embodied tasks, PaLM-E~\cite{driess2023palm} as an embodied model tries to transfer knowledge from visual-language into embodied reasoning tasks and achieves state-of-the-art performance on three different robot embodiments and several general visual-language tasks. 
Although several efforts have been made to explore LLMs in embodied AI, there are still three concerns in knowledge engineering for current scene knowledge research. 
Firstly, current foundation models only have some abilities of probability-driven knowledge base rather than knowledge engineering in the strict sense. The plausible knowledge sorted in the learnable parameters is fuzzy and exists uncontrollable uncertainty, which will be fatal to physical environment reasoning and decision if incorrect knowledge is injected into embodied models. 
Secondly, LLMs do not consider linking the real world directly during training data organization and training process, which requires high-quality prompting engineering and debiases operation to match embodied task requirements~\cite{bian2023chatgpt}. They cannot effectively distinguish between relevant and irrelevant knowledge and generate high-noise knowledge frequently. 
Thirdly, parameter knowledge is a black box compared to loosely-coupled knowledge following knowledge engineering technology. For the former, knowledge is hard to maintain and update dynamically, and the effectiveness of knowledge can only evaluate from the task performance missing ability of intermediate intervention. In contrast, the typical knowledge graph is in the form of explicit representation increasing the integrated reliability of knowledge.

Summarily, current scene knowledge acquisition mainly relies on general knowledge methods. For symbolic knowledge, there are schema design, cost and efficiency of data collection, and quality control problems~\cite{xue2022knowledge}. For parameter knowledge, it is inevitable to guarantee the reliability of knowledge. Dynamic knowledge update for large language models is also a huge challenge.
To overcome the above challenges of scene knowledge for embodied AI, we take \syx{an innovative}  way to combine the advantages of symbolic knowledge and parameter knowledge.
We propose a scene-driven multimodal knowledge graph (Scene-MMKG) construction method for a specific scene, shown in Fig.~\ref{fig:overall}. \syx{It integrates perceptual and apperceptive knowledge about all entities in the scene with multimodal properties. The core ideas contain two parts. 
\textbf{Firstly}, we make use of general knowledge in foundation models with prompt engineering to design scene-driven schema. Then, based on the scene-driven schema, apperceptive knowledge is extracted from existing rich knowledge bases and perceptual knowledge is collected from multimodal situated physical data.}
It provides an easy way to design scene-driven schema and reduces the cost of data collection. The scene-driven prompts eliminate irrelevant information collection to guarantee the capacity and quality of the knowledge base. 
\textbf{Secondly}, benefit from Scene-MMKG construction advantages, scene knowledge can use unified knowledge representation for specific embodied tasks, which enhances the connection of scene knowledge and downstream tasks. Our contributions can be summarized as follows.

\begin{enumerate}
    \item We propose a novel scene-driven multimodal knowledge graph construction method, which can not only overcome the limitations of conventional knowledge graph in embodied tasks, but also complement with large language models in scene information, explainability, and practical implementation, as shown in Fig.~\ref{fig:kg_construction}.

    \item An unified scene knowledge injection framework is proposed to improve typical embodied tasks in mobility and manipulation, as shown in Fig.~\ref{fig:kg_enhanced_model}.

    \item An instantiated indoor knowledge graph, named \textbf{ManipMob-MMKG}, is constructed to evaluate the superiority and effectiveness of our proposed method, which also provides a useful knowledge base for embodied AI community. 

    \item Experiments on two typical embodied tasks are performed, in which the results show that ManipMob-MMKG constructed following our proposed scene-driven method achieves broad advantages compared to methods both using conventional general knowledge graph and parameter knowledge from pre-trained models.
\end{enumerate}

\section{Preliminaries}\label{sec:kg_define}
From the perceptive of embodied AI, the external knowledge of an agent required is multimodal and scene-oriented considering the agent's character of interacting with the environment and multimodal perception. The scene-oriented knowledge for an agent is cross-domain and fine-grained to facilitate scene understanding and decision-making tasks. Therefore, the definition of scene-driven multimodal knowledge can be formally given below:

\textit{\textbf{Scene-driven Multimodal Knowledge Graph (Scene-MMKG)} is an explicit conceptualisation to  agent-environment interaction in a specific scenario, represented in terms of semantically interrelated entities and relations with multimodal properties.}    

The conventional knowledge graph (KG) can be usually classified into general knowledge graph and domain-specific knowledge graph according to the application fields.
A general knowledge graph is usually domain-independent and broadly informative, while suffering in domain knowledge and information granularity. Domain-specific knowledge is associated with domain problems, which focus on fine-grained domain information \syx{with limited scales}. In contrast, our proposed scene-driven knowledge graph has two core characteristics, namely: 
\begin{enumerate}
\item {\bf The boundary of knowledge}. The KG only covers scene-oriented knowledge with fine-grained information, not retrieving all superordinate and subordinate-relationships of \syx{entities}. It can constrain the scale of KG to make it easily.

\item {\bf The content of knowledge}. The KG only concerns about agent-environment interaction (the key idea of Embodied AI) referring to situated object perception, event prediction, decision-making, etc. It frames the scope of the knowledge, reducing irrelevant information collection.
\end{enumerate}

\section{Scene-MMKG Construction}\label{sec:construction_method}
To construct a Scene-MMKG, we depict a specific scene by a series of natural scene profiles from human. The pipeline of construction contains three parts: \textit{prompt-based schema}, \textit{knowledge population}, \textit{quality control and refinement}. Firstly, we consider a prompt-based method to construct a scene-oriented schema automatically and effectively. Secondly, based on the existing general knowledge constrained by scene-driven schema and limited-scale multimodal data collection, we can construct a Scene-MMKG easily and flexibly. The third part is to refine the constructed Scene-MMKG to eliminate long-tail problems.


\subsection{Prompt-based Schema Design}\label{sec:schema_design}


\begin{figure*}[!t]
\centering
\includegraphics[width=0.8\linewidth]{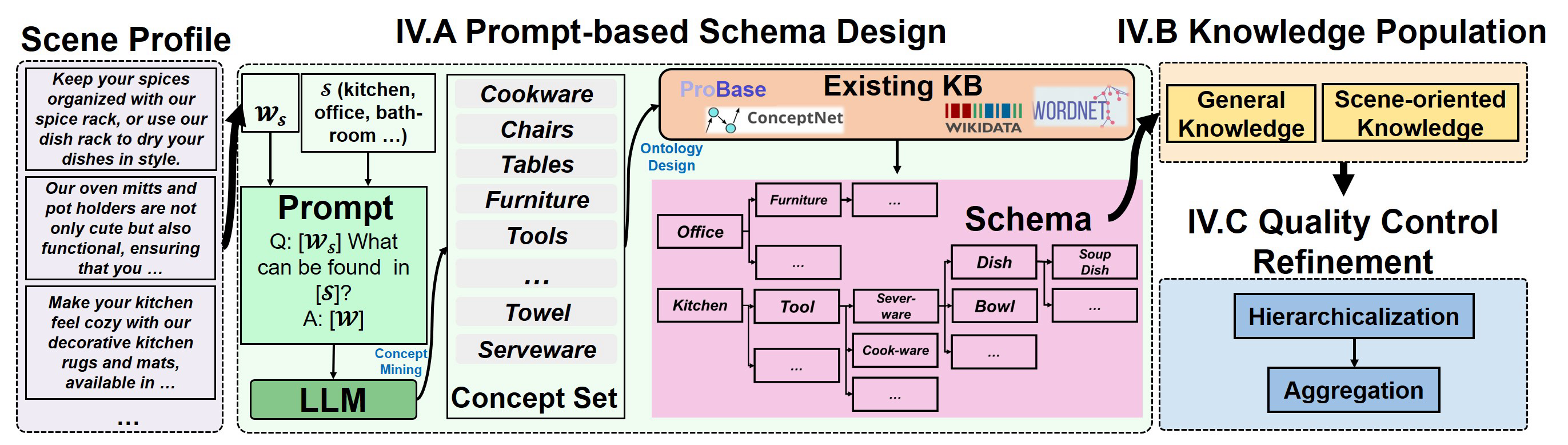}
\caption{Given the scene profiles, we design a prompt-based schema based on LLMs and then populate multimodal knowledge guided by the schema to construct our Scene-MMKG. Scene-MMKG is refined by hierarchicalization and aggregation for attributes to resolve long-tail problems.}
\label{fig:kg_construction}
\end{figure*}

The core characteristic of Scene-MMKG is scene-oriented property, under which we collect multimodal data to guarantee the boundary and content of scene knowledge. The schema design is considered during the construction primarily, which is a hierarchical ontology to describe the relationships between human-event-stuff in the embodied scenario. The conventional method of scheme design is usually manual and has well-established ontologies and taxonomies in a specific domain (e.g. medicine or finance schema). To obtain the schema out of well-established ones, it requires expert knowledge and experience, which is time-consuming, labor-intensive, and quality-uncontrolled for different scenes. To overcome these flaws in schema design, we introduce a flexible, general, pragmatic schema design method to design a high-quality schema automatically using prompt engineering, shown in Fig.~\ref{fig:kg_construction}.


\subsubsection{Scene Concept Mining}
The first step of our schema design is to extract scene concept set $\mathcal{C}$ from the open world $\mathcal {OW}$. We consider a specific scene $\mathcal S$ (e.g. kitchen, bedroom, and office) by a series of natural scene profiles $\mathcal W_{\mathcal S}=\{w_0, w_1, w_2,...,w_n\}$, which are the textual descriptions about the characteristics of the specific scene. Our method seek to model a probability model to collect scene concepts $\mathcal{C}$:
\begin{equation}\label{eq:joint_model}
\begin{aligned}
P(\mathcal{C}| \mathcal{W_S,S}), \mathcal{C} \subset \mathcal{OW}.
\end{aligned} 
\end{equation}

Specifically, we propose a hard-prompt-based scene concept method using LLM in zero-shot settings. It includes \textit{Prompt Addition} and \textit{Answer Search}.

\noindent \textbf{Prompt Addition}\quad In this part, a \textit{prompt function} $f_{prompt} (\cdot)$ is applied to modify the input $\mathcal{W_S}$ and scene $\mathcal{S}$ into a \textit{scene prompt}=$f_{prompt}(\mathcal{W_S, S})$, called as \textit{scene prompt}. This function consists of a two-step process~\cite{liu2023pre}:

\begin{enumerate}
    \item Apply a template, which is a textual string that has two slots: a scene profile slot [$\mathcal{W_S}$] and a scene slot [$\mathcal S$], which will be later mapped into final answer $\mathcal{C_\text{raw}}$.
    \item Fill slots [$\mathcal{W_S}$] and [$\mathcal S$] with the input text $\mathcal{W_S}$ and $\mathcal S$ respectively, as shown in Fig.~\ref{fig:kg_construction}.
\end{enumerate}

\noindent \textbf{Answer Search}\quad To search a potential scene concepts $\mathcal{C}$, we adopt a pre-trained LLM $P (\cdot; \theta)$ with frozen parameters $\theta$ to instantiate probability model Eq.~\ref{eq:joint_model} in zero-shot setting. The whole pipeline can be formulated as below:
\begin{equation}
\begin{aligned}
& \mathcal{C_\text{raw}}  = \underset{\mathcal{C_\text{raw}} \subset \mathcal OW}{search}    \; P(f_{prompt}(\mathcal{W_S, S});\theta),
\end{aligned} 
\end{equation}
where search function $search(\cdot)$ could be an $argmax$ search that searches for the highest-scoring output or sampling that randomly generates outputs following the probability distribution of the adopted LLM. All prompts based on different scene profiles $\mathcal{W_S}$ are fed into the LLM and the outputs are organized as the raw concept set $\mathcal{C_\text{raw}}$.

\subsubsection{Ontology Design}
To obtain the abstract domain concept of schema, we use concept expansion and concept cluster. 

\noindent \textbf{Concept Expansion}\quad  Most of concepts in the raw concept set $\mathcal{C_\text{raw}}$ are concrete or incomplete about a specific domain. Therefore, we propose the concept expansion to augment scene concept in abstraction and completeness (Eq.~\ref{eq:concept_expansion}) using hypernyms and hyponyms in existing knowledge graphs $\mathcal{K}$, such as Probase~\cite{wu2012probase} and WordNet~\cite{miller1995wordnet}. We consider the expanded scene concept $\mathcal{C_{\exp}}$ based on concrete scene concept set $\mathcal{C_\text{raw}}$.
\begin{equation}\label{eq:concept_expansion}
\begin{aligned}
\mathcal{C_{\exp}} = \{ c, k|k \in R_{\text{hypo}}(c) \cup R_{\text{hyper}}(c)\}_{k \in \mathcal{K},c \in \mathcal{C_\text{raw}}},
\end{aligned} 
\end{equation}
where $R_{\text{hypo}}(c)$ and $ R_{\text{hyper}}(c)$ are the functions to extract hypernyms and hyponyms from existing knowledge graphs. The concepts $k$ is added in the expanded scene concept set, which include more abstract concepts from hypernyms and missing concepts from hyponyms. For instance, we expand ``\bluecallig{disposable chopsticks}" to include related hypernyms such as ``\bluecallig{tableware}", ``\bluecallig{disposable cutlery}", and their hyponyms such as ``\bluecallig{spoon}", ``\bluecallig{flatware}”, ``\bluecallig{disposable bowl}" and ``\bluecallig{plastic tablecloth}" if searching for raw scene concept ``\bluecallig{disposable chopsticks}". The search range is usually no more than two layers for hypernyms and hyponyms.

\noindent \textbf{Concept Cluster}\quad  Expanded scene concept $\mathcal{C_{\exp}}$ exists overlapping semantic concepts which leads to abundant in schema design. We aggregate a group of concepts in expanded scene concept $\mathcal{C_{\exp}}$ based on semantic similarity to generate the final scene concept set $\mathcal{C}$ (i.e. \textbf{scene-driven schema}) following Eq.~\ref{eq:concept_cluster}.
\begin{equation}\label{eq:concept_cluster}
\begin{aligned}
&\forall c_{i}, c_{j} \in \mathcal{C_\text{exp}}, c_{i}\neq c_{j},\\
\quad & \mathcal{C}  = 
\begin{cases} 
merge(c_{i}, c_{j}) & \text{if } sim(c_{i}, c_{j}) \geq \gamma_1 \\
c_{i}, c_{j} & \text{otherwise}
\end{cases},
\end{aligned} 
\end{equation}
where $\gamma_1$ is a threshold value for the semantic similarity and the default value is $0.7$.
$sim(\cdot, \cdot)$ is the function that measures the semantic similarity between two concepts. $merge(\cdot, \cdot)$ is an aggregation function to merge two concepts considering hyponymy.

\subsection{Knowledge Population}
To construct scene-driven multimodal knowledge graph, we collect scene data based on the \textbf{scene-driven schema} mentioned in Sec.~\ref{sec:schema_design}. The embodied scenes are various and flexible which results in expensive and time-consuming data collection and annotation. Furthermore, there exist a number of scenes that are hard to situate and obtain direct information.
Therefore, we propose a two-step knowledge population method using existing general knowledge base and automatically collected scene-oriented knowledge. The former is easy-obtained and could populate a large number of general knowledge, while the latter only requires to collect limited-scale scene-specific knowledge and multimodal information to complete the former. The knowledge population pipeline can be formulated as below:
\begin{equation}
\begin{aligned}
\mathcal{K_\text{scene-KG}}&=f_{fill}( \mathcal{C}, \mathcal{K_\text{g}}), \\
\mathcal{K_\text{scene-MMKG}}&=f_{fill}\left(\mathcal{C}, deconflict( \mathcal{K_\text{scene-KG}},  \mathcal{K_\text{s}}) \right),
\end{aligned}
\label{fill_knowledge} 
\end{equation}
where $f_{fill}$ fills knowledge into our proposed $\mathcal{K_\text{scene-MMKG}}$ under the guidance of scene-driven schema $\mathcal{C}$. $\mathcal{K_\text{g}}$ and $\mathcal{K_\text{s}}$ are the general knowledge and scene-oriented knowledge respectively. $\mathcal{K_\text{scene-KG}}$ is the intermediate knowledge graph constructed using existing general knowledge and
$deconflict(\cdot, \cdot)$ resolves conflicts between two knowledge sources. If there are conflicts between $\mathcal{K_\text{scene-KG}}$ and $\mathcal{K_\text{s}}$, we reserve $\mathcal{K_\text{s}}$.

\subsubsection{General Knowledge}
General knowledge $\mathcal{K_\text{g}}$ refers to the set of common or widely accepted facts and concepts that are not specific to a particular scene.
We retrieve knowledge from existing knowledge base, such as ATOMIC${_{20}^{20}}$~\cite{hwang2021comet}, WebChild~\cite{tandon2014webchild} and ConceptNet~\cite{liu2004conceptnet}, based on scene-driven schema $\mathcal{C}$ and mount retrieved knowledge into our intermediate knowledge graph $\mathcal{K_\text{scene-KG}}$.
The retrieved knowledge involves perceptual knowledge (e.g. physical properties: color, shape, size, etc.) and apperceptive knowledge (e.g. experience and common sense: affordance, maintenance, etc.).
$\mathcal{K_\text{g}}$ is a high-quality and cheap knowledge source for our proposed scene-driven knowledge graph to provide part of general knowledge between different scenarios.
Our schema constrains knowledge boundary effectively to guarantee the limited scale of Scene-MMKG and also populate a large amount of knowledge reducing data collection costs.
However, the accuracy and relevance of knowledge to a specific scene are insufficient and coarse-grained derived from the existing general knowledge base. Most of the existing knowledge base is unimodal (text or language), which lacks perceptible modalities (e.g. vision). It is necessary to complete scene-oriented fine-grained knowledge and multimodal information for a specific scene~\cite{jensen1995specific} to improve the content and boundary of Scene-MMKG.



\subsubsection{Scene-oriented Knowledge}
Scene-oriented knowledge $\mathcal{K_\text{s}}$ is the knowledge relying on the specific scene and cannot be obtained from existing general knowledge base. It is an important part in our proposed Scene-MMKG for embodied tasks. We collect the scene-oriented knowledge from the open world guided following designed scene-driven schema $\mathcal{C}$. It is available to obtain the scene-oriented knowledge from multiple sources (e.g. Internet, somatosensory game~\cite{wang2020gesture}), which could bring internet-scale multimedia data for a specific scene in text, image, video, etc.~\footnote{In current instantiation Scene-MMKG, we explore text-image version initially.} and develop a comprehensive understanding of the scene.


$\mathcal{K_\text{s}}$ usually contains multimodal knowledge in an embodied scenario. Text and vision are two typical modalities for knowledge information. Textual knowledge often refers to the information that existing general knowledge does not contain or describes detailedly. For visual knowledge, most of the existing general knowledge base is unimodal using text data, while multimodal data in the knowledge base scopes still exist data desert issues. Our Scene-MMKG is to collect visual data to alleviate the problems in embodied AI. Also, visual knowledge is even more important than text data since it can aid an embodied agent to make decisions in the real world more effectively. We categorize visual knowledge into synthetic knowledge and real-world knowledge~\cite{su2015render}:

\begin{enumerate}
\item Synthetic knowledge: The images are usually rendered by CAD software typically featuring a white or transparent background. They provide the entity's naive appearance, shape, size, color, and so on, while do not feature entity-related scenarios.

\item Real-world knowledge: The images are captured or rendered in a \syx{situated scenario}, showcasing the whole entity with its implemented surroundings, which not only contains various entity-related proprieties but also provides potential scene context. It is hard to obtain completely compared to synthetic knowledge.
\end{enumerate}

\begin{algorithm}[tp]
\small
    \SetAlgoLined 
	\caption{Attribute Hierarchicalization}
        \label{alg:hierarchy}
	\KwIn{one entity $e \in \mathcal E$ attribute set $\mathcal{A}$}
	\KwOut{Hierarchical attribute set $\mathcal{A_\text{hiera}}$}
        
	$\mathcal{A_\text{hiera}} = \emptyset$\; 
        \For{$a_{i}$ in $\mathcal{A}$}{
            \eIf{$a_{i}$ can be subdivided}{
                $P_{i}, AP_{i} = \text{subdivide}(a_{i})$\;
                
        	\For{$p_{ij}$ in $P_{i}$}{
                    $attribute = link(p_{ij}, A)$ \;
                    $\mathcal{A_\text{hiera}}.add(attribute)$ \;
                    \For{$ap_{ik}$ in $AP_{i}$}{
                        \If{$ap_{ik}$ belongs to $p_{ij}$}{
                            $attribute = link(ap_{ik}, p_{ij})$ \;
                            $\mathcal{A_\text{hiera}}.add(attribute)$ \;

        		      }
                    }
        	}
            }
            {
            $\mathcal{A_\text{hiera}}.add(a_{i})$
            }
        }
\end{algorithm}

\subsection{Quality Control and Refinement}
It is inevitable for attributes in our proposed scene-driven multimodal knowledge graph $\mathcal{K_\text{scene-MMKG}}$ to have long-tail distribution. For instance, the instantiated indoor Scene-MMKG comprises $1,580$ attributes, of which $75 \%$ are concentrated within the top $600$ attributes and many attributes are long-tail. It complicates the process of generating knowledge embedding and knowledge injection into downstream tasks~\cite{wang2021reform,zhang2019long}.

These two major causes for long-tail distribution:
\begin{enumerate}
\item Some attributes exhibit composite meaning. 
For instance, ``\bluecallig{frame length}" encompasses both ``\bluecallig{frame}" and ``\bluecallig{length}".

\item There are overlapping attributes from different domains. 
For instance, ``\bluecallig{measure}" and ``\bluecallig{size}" from different knowledge sources could be considered equivalent attributes.
\end{enumerate}
\syx{To address the causes, we propose Quality Control and Refinement (QC\&R) to disambiguate polysemous attributes into distinct, unambiguous attributes through attribute hierarchicalization.}
Hierarchical attributes are Aggregation according to their semantics to deal with pseudo-long-tail problems.

\noindent \textbf{Hierarchicalization}\quad We organize attributes considering the constituent parts of the entity. Compared to linking the \textit{[part + attribute]} to the entity directly, we decompose the entities into constituent parts and mount general attributes to parts. Parts as constituent attribute is mounted to the entity.
The pipeline can be formulated in Algorithm~\ref{alg:hierarchy}, where $\mathcal{A}$ denotes a set of attributes for one entity $e \in \mathcal E$ in our knowledge graph. $a_{i}$ is one attribute of $\mathcal{A}$. If $a_{i}$ can be \textbf{\textit{subdivided}} into some parts $\mathcal{P_\text{i}}$ and some general attributes $AP_{i}$, each element of part set $\mathcal{P_\text{i}}$ will be \textbf{\textit{linked}} to the entity $e$.
For each general attribute $ap_{ik}$ in $AP_{i}$, if it belongs to one part $p_{ij}$, it will be linked to the part $p_{ij}$. Both two-level attributes are \textbf{\textit{added}} into a new hierarchical set $\mathcal{A_\text{hiera}}$ for the entity $e$.

For instance, a ``\bluecallig{chair}" possesses attributes ``\bluecallig{frame length}" and ``\bluecallig{foot length}". 
To manage these attributes effectively, we subdivide the ``\bluecallig{chair}" into its constituent parts, ``\bluecallig{foot}" and ``\bluecallig{frame}", considering ``\bluecallig{length}" as a general attribute of each part. 
Attributes that cannot be further subdivided, such as ``\bluecallig{usage}" and ``\bluecallig{conservation measures}", are directly associated with the ``\bluecallig{chair}".



\noindent \textbf{Aggregation}\quad By employing pre-trained language models, we compute the similarity between each pair of attributes in $\mathcal{A_\text{hiera}}$, aggregating attributes with similar semantics. The pipeline can be formulated as below:
\begin{equation}
\begin{aligned}
& \forall a_{i}, a_{j} \in \mathcal{A_\text{hiera}}, a_{i}\neq a_{j}\\
\quad & \mathcal{A_\text{hiera}} = 
\begin{cases} 
aggregate(a_{i}, a_{j}) & \text{if } sim(a_{i}, a_{j}) \geq \gamma_2 \\
a_{i}, a_{j} & \text{otherwise}
\end{cases},
\end{aligned} 
\end{equation}
where $\gamma_2$ is the pre-defined threshold (default value is $0.7$) and $aggregate(\cdot, \cdot)$ that aggregates two attributes.
For instance, we treat ``\bluecallig{measurement}" and ``\bluecallig{size}" as identical attributes in embodied scenario. 
This approach relieves long-tail attributes into meaningful clusters, effectively reducing the imbalance of knowledge derived from different sources.


\section{Knowledge-enhanced Embodied Task}
From the robotic perspective, to interact with real-world environment, an embodied agent is equipped with at least one of the functionalities of \textbf{mobility} and \textbf{manipulation}~\cite{khatib1999robots}. It derives a series of specific tasks for embodied intelligence.

\noindent \textbf{Mobility}
Mobility (locomotion) is the ability of an agent to move around in the environment with wheels, crawlers, legs, or propellers~\cite{siciliano2008springer}. Different from map-dependence tasks, embodied task is usually required to move and interact with the environment actively to achieve the goal given ego-centric sensory inputs and referred information, such as visual-language navigation. One of the existing challenges is how to obtain enough scene understanding and encode observed scene under first-person partial perspective. 

\noindent \textbf{Manipulation}
Manipulation refers to an agent's control of its environment through selective contact~\cite{mason2001mechanics}. An agent can operate objects in the environment using arms, end-effectors, humanoid hands, etc. Typical manipulation actions include grasping, pushing, poking, and pick-and-place. One of the existing challenges is how to model fine-grained, dexterous and safety operation on the object with partial or occluded observation.

To evaluate the effectiveness of our proposed scene-driven knowledge graph for embodied tasks, we introduce two typical tasks (i.e. visual language navigation and 3D object language grounding) corresponding to mobility and manipulation.

\begin{figure*}[!t]
\centering
\vspace{-0.3cm}  
\includegraphics[width=0.8\linewidth]{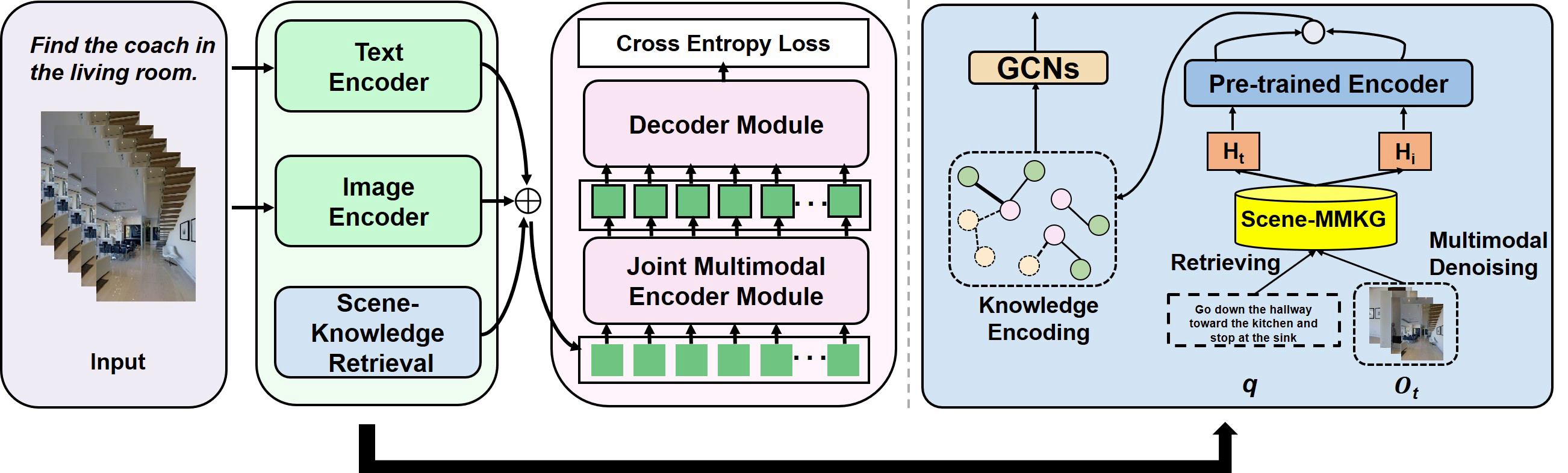}
\caption{The overview of the scene-driven knowledge enhancement model is shown in the left panel. The right panel is the details about scene knowledge retrieval module.}
\label{fig:kg_enhanced_model}
\end{figure*}

\subsection{Scene-driven Knowledge Enhancement Model}\label{sec:gate}
To address above mentioned challenges in embodied tasks (i.e. mobility and manipulation), we propose a knowledge-enhanced method, as shown in Fig.~\ref{fig:kg_enhanced_model}. 

We design the \textit{Scene Knowledge Retrieval (SKR)} module to retrieve scene knowledge from the Scene-MMKG. It includes \textit{retrieving}, \textit{multimodal denoising}, and \textit{knowledge encoding} operations.

\noindent\textbf{Retrieving} The agents make decisions based on one instruction, and the varying observation images for each time step.
Therefore, SKR retrieves knowledge $\hat{e}$ based on the similarity of query from observation images or instructions $q$ and entities $\mathcal E$ in Scene-MMKG following Eq.~\ref{eq:retrieving}.
\begin{equation}\label{eq:retrieving}
\begin{aligned}
\hat{e} = \underset{e \in \mathcal E}{\mathrm{argmax}}  \; sim(q, e) ,
\end{aligned} 
\end{equation}
where $sim(\cdot, \cdot)$ calculates the semantic similarity between $q$ and the entity $e$ from $\mathcal E$.
This allows the SKR to find relevant information and provide accurate responses to queries.
The retrieved knowledge set $\hat{e} \in \hat{\mathcal E}$ can be organized as multimodal symbolic triple $H = (H_{t}, H_{v})$, in which $H_{t}$ is the traditional textual knowledge node while $H_{v}$ is the visual knowledge (multimodal) node.

\noindent\textbf{Multimodal Denoising} However, since the retrieved $H$ may contain noise data, especially multimodal knowledge, we introduce a \textit{Multimodal Denoising module} to refine the retrieved knowledge. 
The agent takes an observation $O_t$ at each time step while \textit{Multimodal Denoising module} uses $O_t$ as the query to search the most relevant information from $H_{v}$ and discard irrelevant or noisy data.
\textit{Multimodal Denoising module} calculates the similarity between $O_t$ and $H_{v}$, and selects the $H_{v}$ of which similarity is greater than a pre-defined threshold $\gamma_3$. The process is formulated as below:
\begin{equation}
\begin{aligned}
 H_i = 
\begin{cases}
  & H_i \ \ \ \ \ \text{ if } \ \ \ sim(O_t, H_i)\ge \gamma_3   \\
  & 0   \ \ \ \ \ \ \ \text{ otherwise } 
\end{cases}.
\end{aligned} 
\end{equation}

\noindent \textbf{Knowledge Encoding} We mount the multimodal knowledge $H_{v}$ as a tail node to the symbolic triples.
Each node in graph is encoded using CLIP text encoder and image encoder.
We encode scene knowledge through $n$-layers GCNs~\cite{ebisu2019generalized,fang2022contrastive,liang2023knowledge} and output the knowledge representation $F_{H}$ in the last layer, which can be formulated as below:
\begin{equation} 
\begin{aligned}
H^{0} & = (H_{t}, \sigma (H_{v})), \\
H^{m + 1} & = GCN^{m}(H^{m}), \\
...\\
F_{H} & = H^{n},
\end{aligned} 
\end{equation}
where $GCN^{m}$ is the learnable GCNs at the $m$-th layer, $\sigma(\cdot)$ is the \textit{Multimodal Denoising module},
and $H^{m}$ is the feature matrix of the node after encoding by $m$-th GCN layer.

\subsection{Knowledge Injection in VLN}\label{sec:kg_vln_injection}
Visual Language Navigation (VLN)~\cite{anderson2018vision,deitke2022retrospectives} is a typical task in robot mobility. 
In VLN tasks, the agent is equipped with an ego-centric visual perspective of a realistic, previously unseen environment and tasked with navigating based on a path described in natural language instructions such as ``\textit{Go down the hallway toward the kitchen and stop at the sink}".

The agent detects objects with Faster R-CNN pre-trained on Visual Genome~\cite{krishna2017visual} and encodes objects into $F_{O_{i}}^{t}, i \in [0, n)$, where $n$ is the number of objects that could be found by the agent at step $t$. 
We borrow ideas from Transformer~\cite{vaswani2017attention} and apply them to the sequential decision-making in VLN.
The language features $F_{I}$ is from Transformer encoder based on instructions $I$. 
The scene memory token $m^{t}$ at step $t$ is constructed by concatenation,
where $F_H^{t}$ is the encoded scene knowledge from Scene-driven Knowledge Enhancement Module at step $t$.
After that, Transformer \textit{Decoder} takes history scene memory tokens ${m^{i}, t \in [0, t)}$ as inputs to produce a hidden state $h^t$:
\begin{equation}
\begin{aligned}
m^{t} & = [F_{I}, F_{O_{i}}^{t}, F_H^{t}], \\
h^{t} & = Decoder([m^{0}, m^{1}, ..., m^{t}]).
\end{aligned} 
\end{equation}
We train with cross-entropy-loss $\mathcal{L}_{s} (a^{t}, a_{t}^{*}) $ to predict each step action $a_{t}$, where $a_{t}^{*}$ is the ground truth action. $a_{t}$ is the predicted action based on hidden state $h^t$ with softmax function.

\subsection{Knowledge Injection in 3D Object Language Grounding}\label{sec:kg_grd_injection}
3D Object Language Grounding~\cite{thomason2022language} refers to agents identifying objects based on color, shape, and category by modeling the relationships between language and physical objects.
It is the preliminary step of manipulating objects using appliances and tools, which can link the virtual cognitive world and the real world.

We use the CLIP image encoder to extract the image features $F_{O_{A}}$ and $F_{O_{B}}$ which are max-pooled from observation images of two objects $O_{A}$ and $O_{B}$. CLIP text encoder is used to extract the sentence features $F_{I}$ from sentence $I$ similar to the pipeline of MATCH~\cite{thomason2022language}. 
$F_{H}$ is the scene knowledge from Scene-driven Knowledge Enhancement Module based on sentence $I$. 
The CLIP encoder is pre-trained and we freeze the parameters of encoder during training. We concatenate these features to extract multimodal features $s_A$ and $s_B$ using multi-layer perceptron (MLP). Then, we concatenate $s_A$ and $s_B$ to predicate a score $s$ with MLP. 
\begin{equation}
\begin{aligned}
 s_{A} & = MLP([F_{I}, F_{O_{A}}, F_H]), \\
 s_{B} & = MLP([F_{I}, F_{O_{B}}, F_H]), \\
 s & = MLP([s_{A}, s_{B}]),
\end{aligned} 
\end{equation}
where $MLP$ is the multi-layer perceptron. 
We train with cross-entropy-loss $\mathcal{L}_{s} (s, s^{*}) $ to predict a binary label of whether the referring expression matches the object, where $s^{*}$ is the ground truth object.

\begin{table*}[t]
\centering
\vspace{-0.3cm}  
\caption{\syx{Characteristics of existing knowledge used in embodied tasks.}}
\resizebox{0.88\linewidth}{!}{

\begin{tabular}{@{}cc|c|c|cc@{}}
\toprule
\multicolumn{2}{c|}{\multirow{2}{*}{Knowledge}}                                                                                                                               & \multirow{2}{*}{Cost}                                                                                                                                         & \multirow{2}{*}{Parameters}                                                                                     & \multicolumn{2}{c}{Scale}                                                                                                                                                                                                                                                                                                                  \\ \cmidrule(l){5-6} 
\multicolumn{2}{c|}{}                                                                                                                                                         &                                                                                                                                                               &                                                                                                                 & \multicolumn{1}{c|}{Global Scale}                                                                                                               & Scale using in Task                                                                                                                                                                      \\ \midrule
\multicolumn{1}{c|}{\multirow{3}{*}{\begin{tabular}[c]{@{}c@{}}Implicit\\ Knowledge\end{tabular}}} & ViLBERT~\cite{lu2019vilbert}                                                                  & 8 NVIDIA TitanX for 10 epochs                                                                                                                                 & 268 million total parameters                                                                                    & \multicolumn{1}{c|}{-}                                                                                                                          & -                                                                                                                                                                                        \\ \cmidrule(l){2-6} 
\multicolumn{1}{c|}{}                                                                              & LegoFormer~\cite{yagubbayli2021legoformer}                                                               & \begin{tabular}[c]{@{}c@{}}NVIDIA A100 and T4 GPUs on Google Cloud\\ Platform (GCP) for 160K steps (LegoFormer-M) \\ or 80K steps (LegoFormer-S)\end{tabular} & 168 million total parameters                                                                                    & \multicolumn{1}{c|}{-}                                                                                                                          & -                                                                                                                                                                                        \\ \cmidrule(l){2-6} 
\multicolumn{1}{c|}{}                                                                              & BLIP-2~\cite{li2023blip}                                                                   & \begin{tabular}[c]{@{}c@{}}16 NVIDIA A100 for 250K steps (6 Days)\\ in the first stage and 80K steps (3 Days)\\ in the second stage\end{tabular}              & \begin{tabular}[c]{@{}c@{}}12.1 billion total parameters\\ and 108 million training parameters\end{tabular}     & \multicolumn{1}{c|}{-}                                                                                                                          & -                                                                                                                                                                                        \\ \midrule
\multicolumn{1}{c|}{\multirow{5}{*}{\begin{tabular}[c]{@{}c@{}}Explicit\\ Knowledge\end{tabular}}} & ConceptNet~\cite{liu2004conceptnet}                                                               & \begin{tabular}[c]{@{}c@{}}1,4000 contributors of OMCS web project\\ for 4 years and over 3,000\\ volunteers for evaluation\end{tabular}                      & -                                                                                                               & \multicolumn{1}{c|}{\begin{tabular}[c]{@{}c@{}}21 million edges and \\ over 8 million nodes\end{tabular}}                                       & 1,600 entities and 23,306 edges for VLN                                                                                                                                                  \\ \cmidrule(l){2-6} 
\multicolumn{1}{c|}{}                                                                              & LINK~\cite{liu2023learning}                                                                     & \begin{tabular}[c]{@{}c@{}}Two sets of AMT workers with \\ extra 3 annotators for quality evaluation\end{tabular}                                             & -                                                                                                               & \multicolumn{1}{c|}{\begin{tabular}[c]{@{}c@{}}1,457 situated objects, 15 multimodal \\ properties types and 200 total properties\end{tabular}} & -                                                                                                                                                                                        \\ \cmidrule(l){2-6} 
\multicolumn{1}{c|}{}                                                                              & \begin{tabular}[c]{@{}c@{}}ManipMob-MMKG\\ (w/o QC\&R)\end{tabular}      & \multirow{3}{*}{\begin{tabular}[c]{@{}c@{}}5 volunteers for the knowledge population and \\ refinement about 4 days\end{tabular}}                             & \multirow{3}{*}{\begin{tabular}[c]{@{}c@{}}3.1 million parameters in\\ knowledge retrieval module\end{tabular}} & \multicolumn{1}{c|}{\begin{tabular}[c]{@{}c@{}}\textbf{261,440} edges, 11,750 nodes\\ and 34,896 images\end{tabular}}                                    & \begin{tabular}[c]{@{}c@{}}1,600 entities, \textbf{35,410} edges\\ and 10,186 images for VLN;\\ 40,641 entities, \textbf{545,094} edges\\ and 161,167 images for 3D object language grounding\end{tabular} \\ \cmidrule(lr){2-2} \cmidrule(l){5-6} 
\multicolumn{1}{c|}{}                                                                              & \begin{tabular}[c]{@{}c@{}}ManipMob-MMKG\\ (scene-related)\end{tabular}  &                                                                                                                                                               &                                                                                                                 & \multicolumn{1}{c|}{\begin{tabular}[c]{@{}c@{}}\textbf{82,820} edges, 11,750 nodes\\ and 34,896 images\end{tabular}}                                     & \begin{tabular}[c]{@{}c@{}}1,600 entities, \textbf{11,213} edges\\ and 10,186 images for VLN;\\ 40,641 entities, \textbf{214,497} edges\\ and 161,167 images for 3D object language grounding\end{tabular} \\ \cmidrule(lr){2-2} \cmidrule(l){5-6} 
\multicolumn{1}{c|}{}                                                                              & \textbf{\begin{tabular}[c]{@{}c@{}}ManipMob-MMKG \\ (ours)\end{tabular}} &                                                                                                                                                               &                                                                                                                 & \multicolumn{1}{c|}{\begin{tabular}[c]{@{}c@{}}\textbf{178,620} edges, 11,750 nodes\\ and 34,896 images\end{tabular}}                                    & \begin{tabular}[c]{@{}c@{}}1,600 entities, \textbf{28,684} edges\\ and 10,186 images for VLN;\\ 40,641 entities, \textbf{515,302} edges\\ and 161,167 images for 3D object language grounding\end{tabular} \\ \bottomrule
\end{tabular}
}
\label{tb:main_comparison_of_kb}
\end{table*}

\section{Instantiation of Scene-MMKG}
To evaluate the advantages of our scene-driven multimodal knowledge graph (Scene-MMKG), we instantiate a Scene-MMKG considering typical indoor robotic functionalities (i.e.\textbf{ Manip}ulation and \textbf{Mob}ility), named \textbf{ManipMob-MMKG}, which is a representative and complex scenario for a service robot. The characteristics of ManipMob-MMKG can be seen in TABLE~\ref{tb:main_comparison_of_kb}. We analyze ManipMob-MMKG guided by our proposed knowledge graph construction method in Sec.~\ref{sec:construction_method} from \textit{Data Efficiency} and \textit{Scale and Granularity}.

\subsection{Data Efficiency}\label{efficiency}
The construction of a knowledge base is a labor-intensive and costly task. 
According to Paulheim's evaluation~\cite{paulheim2018much}, the cost of manual knowledge graph creation ranges from over $\$2$ per statement for CYC~\cite{lenat1995cyc} and Freebase~\cite{bollacker2008freebase}, to about $\$0.15$ per statement for projects like DBpedia~\cite{lehmann2015dbpedia} and NELL~\cite{mitchell2018never}. 
This cost increases with the scale of the knowledge base. We compare the construction costs of different knowledge bases adopted by embodied tasks in Sec.~\ref{sec:experiment}.

The construction of explicit knowledge requires data collection and annotation, which requires significant human effort. 
For instance, as shown in TABLE~\ref{tb:main_comparison_of_kb}, ConceptNet~\cite{liu2004conceptnet} project originated from the Open Mind Common Sense project, which involved over 1,400 contributors. \syx{LINK~\cite{liu2023learning} employed two sets of AMT workers for data collection and extra annotators for quality evaluations.}
Implicit knowledge, on the other hand, requires GPU training for several days. 
For instance, the ViLBERT~\cite{lu2019vilbert} model required at least 8 NVIDIA TitanX GPUs, and BLIP-2~\cite{li2023blip} took about 9 days to train for two stages with frozen parameters of Vision Transformers (ViTs) and LLMs. In addition, for 3D vision knowledge, LegoFormer~\cite{yagubbayli2021legoformer} needs NVIDIA A100 and T4 GPUs for pre-training.
In contrast, our ManipMob-MMKG only needed 5 volunteers for 4 days to construct independent of high-performance computation resources. Explicit knowledge requires data collection and annotation, while implicit knowledge requires significant computational resources. Our proposed scene-driven knowledge could keep the balance cost between computation and labor consumption.





\subsection{Scale and Granularity}
For explicit knowledge, the scale of the knowledge base is significant for downstream embodied tasks. 
As shown in TABLE~\ref{tb:main_comparison_of_kb}, ConceptNet contains 21 million nodes and 8 million edges, \syx{while LINK consists of $1,457$ object instances associated with 15 types of 200 multimodal properties (only $208$ images)}.
In contrast, our ManipMob-MMKG only has 11,750 nodes and 178,620 edges, which is relatively smaller than ConceptNet to construct easily. 
However, ManipMob-MMKG includes 34,896 images as multimodal properties that can provide more various external knowledge for embodied tasks. 

Especially, for the visual language navigation (VLN) task, following baseline work~\cite{gao2021room}, we sampled sub-graphs as input to our models from ConceptNet and ManipMob-MMKG respectively. 
For baseline using ConceptNet, we sampled a sub-graph based on 1,600 entities with 23,306 edges that could reproduce the VLN performance same as baseline work~\cite{gao2021room} reported. 
For our ManipMob-MMKG-based model, the sub-graph includes 1,600 entities with 28,684 edges, attaching 10,186 images (indexed by entities) as multimodal knowledge, which could help an agent more comprehensively understand the scenario and effectively decision-making. 
ManipMob-MMKG provides more fine-grained explicit multimodal scene knowledge in a limited scale.



\section{Experimental Results and Analysis}\label{sec:experiment}
We conduct two typical embodied task experiments (mentioned in Sec.~\ref{sec:kg_vln_injection} and Sec.~\ref{sec:kg_grd_injection}) to investigate four research questions (\textbf{RQ}s) about our proposed scene-driven knowledge graph using instantiation \textbf{ManipMob-MMKG}. The first three are about the effectiveness of our proposed scene-driven knowledge graph for embodied tasks. The last one is concerning knowledge selection for embodied tasks.


\noindent\textbf{RQ1:} Is external knowledge important for embodied tasks?

\noindent\textbf{RQ2:} \syx{Does scene-driven knowledge proposed in Sec.~\ref{sec:kg_define} have more advantages than existing general or domain-specific knowledge?} 

\noindent\textbf{RQ3:} How do unimodal knowledge and multimodal knowledge perform on embodied tasks, and which is better?

\noindent\textbf{RQ4:} For fine-grained and widely informative scene knowledge, is it necessary to knowledge-selecting policy? and how to do it?




 

\begin{table*}[htp]
\centering
\vspace{-0.3cm}  
\caption{\syx{Comparisons of visual language navigation on different methods.}}
\resizebox{0.9\linewidth}{!}{
\begin{tabular}{@{}ccc|cccc|cccc|cccc@{}}
\toprule
\multicolumn{3}{c|}{\multirow{2}{*}{Method}}                                                                                                                                                                                                   & \multicolumn{4}{c|}{Val-Seen}                                                                                  & \multicolumn{4}{c|}{Val-Unseen}                                                                                & \multicolumn{4}{c}{Test-Unseen}                                                      \\ \cmidrule(l){4-15} 
\multicolumn{3}{c|}{}                                                                                                                                                                                                                          & \multicolumn{1}{c|}{SR$\uparrow$}   & \multicolumn{1}{c|}{OSR$\uparrow$}  & \multicolumn{1}{c|}{TL$\downarrow$}   & SPL$\uparrow$                        & \multicolumn{1}{c|}{SR$\uparrow$}   & \multicolumn{1}{c|}{OSR$\uparrow$}  & \multicolumn{1}{c|}{TL$\downarrow$}   & SPL$\uparrow$                        & \multicolumn{1}{c|}{SR$\uparrow$} & \multicolumn{1}{c|}{OSR$\uparrow$} & \multicolumn{1}{c|}{TL$\downarrow$} & SPL$\uparrow$   \\ \midrule
\multicolumn{2}{c|}{\multirow{4}{*}{w/o knowledge}}                                                                                                                                     & Random                                               & 2.74                      & 8.92                      & 11.99                     & 1.91                       & 1.76                      & 11.91                     & 10.76                     & 1.01                       & 2.30                    & 8.88                     & 10.34                   & 1.44  \\ \cmidrule(l){3-15} 
\multicolumn{2}{c|}{}                                                                                                                                                                   & FAST-Short~\cite{ke2019tactical}                                           & 45.12                     & 49.68                     & 13.22                     & 40.18                      & 10.08                     & 20.48                     & 29.70                     & 6.17                       & 14.18                   & 23.26                    & 30.69                   & 8.74  \\ \cmidrule(l){3-15} 
\multicolumn{2}{c|}{}                                                                                                                                                                   & FAST-Lan-Only~\cite{ke2019tactical}                                        & 8.36                      & 23.61                     & 49.43                     & 3.67                       & 9.37                      & 29.76                     & 45.03                     & 3.65                       & 8.15                    & 28.45                    & 46.19                   & 2.88  \\ \cmidrule(l){3-15} 
\multicolumn{2}{c|}{}                                                                                                                                                                   & REVERIE~\cite{qi2020reverie}+FAST                                         & 50.53                     & 55.17                     & 16.35                     & 45.50                      & 14.40                     & 28.20                     & 45.28                     & 7.19                       & 19.88                   & 30.63                    & 39.05                   & 11.61 \\ \midrule
\multicolumn{1}{c|}{\multirow{7}{*}{w/ knowledge}} & \multicolumn{1}{c|}{General Knowledge}                                                                                             & CKR-ConceptNet~\cite{gao2021room}                                       & 57.27                     & 61.91                     & 12.16                     & 53.57                      & 19.14                     & \textbf{31.44}            & 26.26                     & 11.84                      & 22.00                   & 30.40                    & 22.46                   & 14.25 \\ \cmidrule(l){2-15} 
\multicolumn{1}{c|}{}                              & \multicolumn{1}{c|}{\multirow{2}{*}{Domain-specific Knowledge}}                                                                    & LINK~\cite{liu2023learning}                                                 & \multicolumn{1}{l}{54.64} & \multicolumn{1}{l}{60.19} & \multicolumn{1}{l}{13.63} & \multicolumn{1}{l|}{51.46} & \multicolumn{1}{l}{18.84} & \multicolumn{1}{l}{25.86} & \multicolumn{1}{l}{26.51} & \multicolumn{1}{l|}{13.09} & -                       & -                        & -                       & -     \\ \cmidrule(l){3-15} 
\multicolumn{1}{c|}{}                              & \multicolumn{1}{c|}{}                                                                                                              & \multicolumn{1}{l|}{CKR-ManipMob-MMKG (Scene-related)} & \multicolumn{1}{l}{56.85} & \multicolumn{1}{l}{67.39} & \multicolumn{1}{l}{12.83} & \multicolumn{1}{l|}{52.24} & \multicolumn{1}{l}{18.23} & \multicolumn{1}{l}{28.71} & \multicolumn{1}{l}{23.76} & \multicolumn{1}{l|}{14.03} & -                       & -                        & -                       & -     \\ \cmidrule(l){2-15} 
\multicolumn{1}{c|}{}                              & \multicolumn{1}{c|}{\multirow{4}{*}{\textbf{\begin{tabular}[c]{@{}c@{}}Scene-driven Knowledge\\ (Scene-MMKG, ours)\end{tabular}}}} & \textbf{CKR-ManipMob-MMKG}                           & \textbf{63.88}            & \textbf{68.54}            & \textbf{11.40}            & \textbf{60.49}             & \textbf{20.04}            & 27.71                     & \textbf{20.38}            & \textbf{16.13}             & -                       & -                        & -                       & -     \\ \cmidrule(l){3-15} 
\multicolumn{1}{c|}{}                              & \multicolumn{1}{c|}{}                                                                                                              & CKR-ManipMob-MMKG (no denoise)                       & 58.69                     & 63.73                     & 12.03                     & 55.24                      & 19.26                     & 27.21                     & 22.49                     & 12.50                      & -                       & -                        & -                       & -     \\ \cmidrule(l){3-15} 
\multicolumn{1}{c|}{}                              & \multicolumn{1}{c|}{}                                                                                                              & CKR-ManipMob-MMKG (unimodal)                         & 57.62                     & 63.11                     & 12.75                     & 53.81                      & 20.51                     & 28.50                     & 25.35                     & 13.74                      & -                       & -                        & -                       & -     \\ \cmidrule(l){3-15} 
\multicolumn{1}{c|}{}                              & \multicolumn{1}{c|}{}                                                                                                              & CKR-ManipMob-MMKG (w/o QC\&R)                        & \multicolumn{1}{l}{56.78} & \multicolumn{1}{l}{67.25} & \multicolumn{1}{l}{12.82} & \multicolumn{1}{l|}{52.28} & \multicolumn{1}{l}{17.55} & \multicolumn{1}{l}{26.81} & \multicolumn{1}{l}{22.24} & \multicolumn{1}{l|}{13.94} & -                       & -                        & -                       & -     \\ \bottomrule
\end{tabular}
}
\label{main_comparison}
\end{table*}

\subsection{Implementation Details}
\subsubsection{Visual Language Navigation}
\noindent\textbf{Datasets}
We train and evaluate our proposed method in Sec.\ref{sec:kg_vln_injection} on dataset REVERIE which is spilt into training, validation, and test sets~\cite{qi2020reverie}. 
The training set consists of $59$ scenes and $10,466$ instructions over $2,353$ objects. The validation-seen set (Val-Seen) contains $53$ scenes, $1,371$ instructions, and $428$ objects. 
The validation-unseen set (Val-Unseen) has $10$ unseen scenes, $3,753$ instructions, and $525$ objects that do not appear in the training set. 
The test set consists of $6,292$ instructions involving $834$ objects in $16$ different scenes.  

\noindent\textbf{Training Details}
We introduce a scene-driven knowledge enhancement method (Sec.~\ref{sec:kg_vln_injection}) by replacing general knowledge ConceptNet~\cite{liu2004conceptnet} in existing state-of-the-art (SOTA) work~\cite{gao2021room} with ManipMob-MMKG. The training details are the same as existing work~\cite{gao2021room}.
We employ the Adam optimizer with weight decay $5e-4$, batch size $100$, and learning rate $1e-4$ during the training stage. 
Experiments are implemented on Ubuntu $18.04$ and PyTorch $1.7.1$ with one Intel Core i9-13900K and two NVIDIA RTX A6000 GPUs (the same as below). 

\noindent\textbf{Metrics}  
We treat it as a successful navigation if the agent reaches the appropriate viewpoints and detects the target within $3$ meters.
Four evaluation metrics~\cite{gao2021room} are used in VLN including Success Rate (SR, reported in percentage \%), Success rate weighted by trajectory Path Length (SPL), Oracle Success Rate (OSR, reported in percentage \%), and Trajectory Length (TL). 
SPL is a primary metric to balance SR and TL.

\subsubsection{3D Object Language Grounding}
\textbf{Datasets} 
We train and evaluate our proposed method in Sec.~\ref{sec:kg_grd_injection} on dataset SNARE which is spilt into training, validation, and test sets following existing work~\cite{thomason2022language}. 
SNARE is a benchmark of the agent must choose the correct object given a language reference of the target object with two image candidates. 
Each example has the label of Visual or Non-visual which means the language reference describes whether the visual features such as the name of the object and colors (e.g. “White mug”) or non-visual features such as shapes and parts (e.g. “Handle is semi-circular”). 
The training set consists of $207$ categories, $6,153$ objects, and over $39,104$ references. 
The validation set contains $7$ categories, $371$ objects, and over $2,304$ references. 
The test set consists of $48$ categories, $1,357$ objects, and over $8,751$ references. 


\noindent\textbf{Training Details}  
We propose a scene knowledge injected method in Sec.~\ref{sec:kg_grd_injection}, named \textbf{L}anguage gr\textbf{O}unding s\textbf{C}ene \textbf{K}nowledge \textbf{E}nhancemen\textbf{T} (\textbf{LOCKET}).  
The data split is the same as existing work~\cite{thomason2022language}.
We employ the Adam optimizer with weight decay $5e-4$, batch size $50$, and learning rate $1e-3$ during the training stage. 

\noindent\textbf{Metrics} 
Accuracy (reported in percentage \%) about correct predictions given one language reference and two candidate objects is used to evaluate the proposed method. 
We calculate the accuracy of the agent choosing the correct object in the Visual labels and Non-visual modes respectively.

\subsection{Visual Language Navigation Results Analysis}\label{vln_result_analysis}

\subsubsection{w/ knowledge vs. w/o knowledge}
For \textbf{RQ1}, the experimental results are shown in TABLE~\ref{main_comparison}. 
We can find that methods using injection knowledge have superior performance compared to their counterparts that lack external knowledge, especially in TL and SPL metrics.
As we can see, the best model on the REVERIE benchmark dataset without using knowledge is the REVERIE+FAST model. 
In unseen scenes, the agent in REVERIE+FAST model is constantly trials to seek unseen targets, and TL is $45.28$ in Val-Unseen. In contrast, knowledge-used models (ConceptNet, ManipMob-MMKG) benefit from the external knowledge to achieve better performance in TL ($26.26$ and $20.38$).
In seen scenes, to compare REVERIE+FAST with our CKR-ManipMob-MMKG, we can find our proposed method achieve nearly $15$ improvement in SPL ($45.50 \rightarrow 60.49$). The performance and generalization of models can be effectively improved by external knowledge injection.

\subsubsection{General or Domain-specific vs. Scene-driven}
\syx{For \textbf{RQ2}, our proposed method CKR-ManipMob-MMKG performs better than CKR-ConceptNet using general knowledge and methods using domain-specific knowledge in most results, especially in TL and SPL. In seen scenes, our proposed method improves TL from $51.46$ to $60.49$ and SPL from $12.83$ to $11.40$ compared to domain-specific methods in TABLE~\ref{main_comparison}.
In unseen scenes, our proposed method achieves $20.38$ and $16.13$ in TL and SPL, which has improvements of $5.88$ ($26.26 \rightarrow 20.38$) and $4.29$ ($11.84 \rightarrow 16.13$) than CKR-ConceptNet. The results show the superiority of our proposed scene-driven knowledge to conventional general or domain-specific knowledge in embodied mobility tasks.
Benefiting rich multimodal scene knowledge encoding from the ManipMob-MMKG, our CKR-ManipMob-MMKG method can effectively identify relevant landmarks and navigate to the destination with fewer unnecessary steps (smaller TL values). 
This not only improves the efficiency of navigation planning but also reduces the risk of making mistakes in unseen scenes.}

\begin{figure}
\centering
\vspace{-0.3cm}
\includegraphics[width=\linewidth]{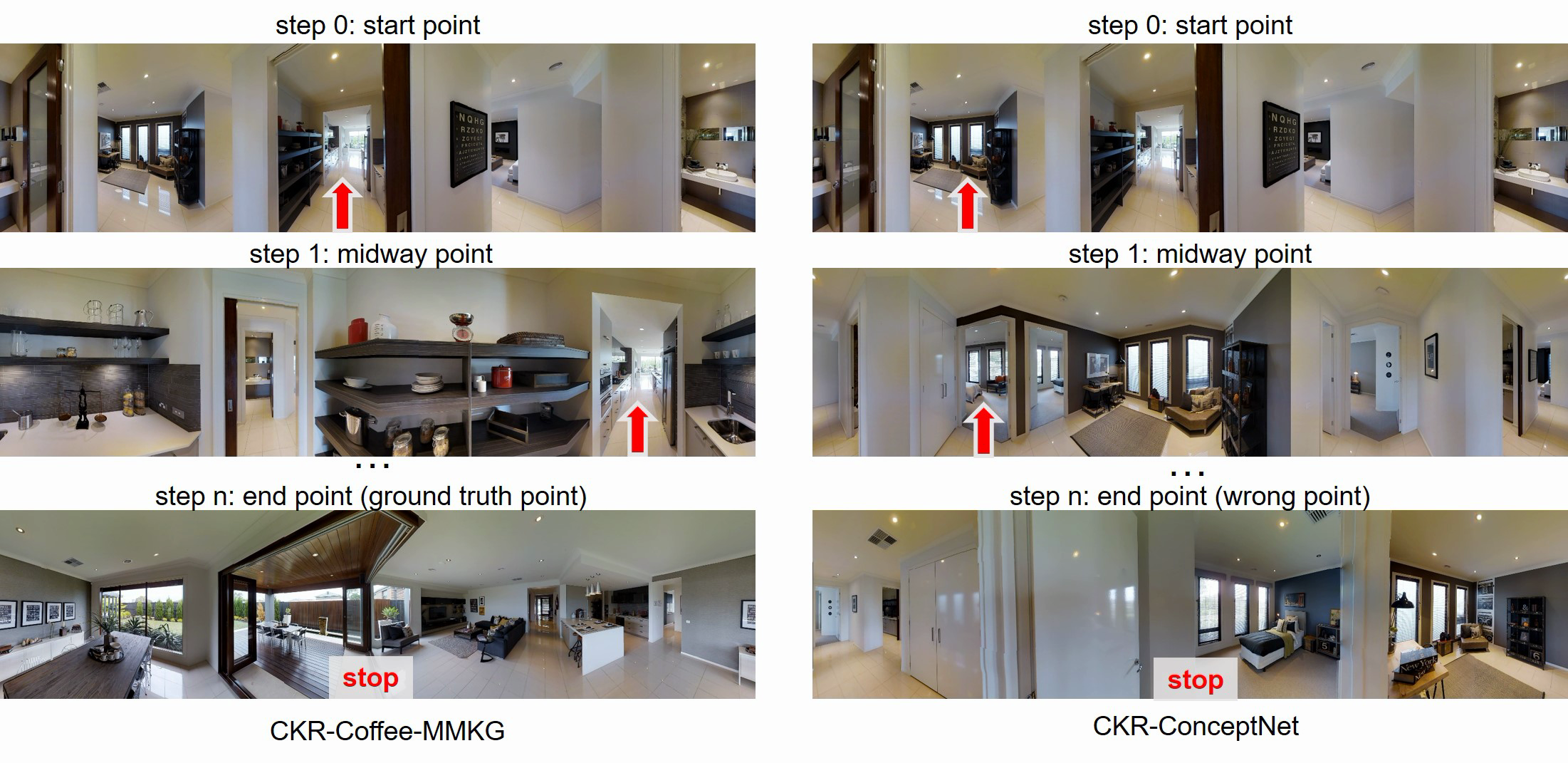}
\caption{Visualization of VLN results of CKR-ManipMob-MMKG (left) and CKR-ConceptNet (right).}
\label{fig:Conceptnet_CKR_MMS_1}
\end{figure}

In Fig.~\ref{fig:Conceptnet_CKR_MMS_1}, we provide visualizations of CKR-ConceptNet and CKR-ManipMob-MMKG results.
As we can see, robot equipped with CKR-ManipMob-MMKG successfully arrives at the intended destination, as shown on the left-hand side following the instruction ``\textit{Go to the dining room and clear everything from the dining table}" (id: $1665\_207\_0$). In contrast, robot equipped with CKR-ConceptNet starts an incorrect trajectories at step $0$ and stops at an incorrect point at step $n$. We attribute this to a misunderstanding of the terms ``\textit{dining room}" and ``\textit{dining table}" guided by general knowledge. 




\subsubsection{Unimodal vs. Multimodal}

\begin{table*}[htp]
\centering
\vspace{-0.3cm}  
\caption{Comparisons of 3D object language grounding on different methods. \syx{Mean accuracy ± standard deviation over random 5 seeds for each method is reported.}}
\resizebox{0.75\linewidth}{!}{
\begin{tabular}{@{}ccc|ccc|ccc@{}}
\toprule
\multicolumn{3}{c|}{\multirow{2}{*}{Method}}                                                                                                                                                        & \multicolumn{3}{c|}{Val}                                        & \multicolumn{3}{c}{Test}                                     \\ \cmidrule(l){4-9} 
\multicolumn{3}{c|}{}                                                                                                                                                                               & All                 & Non-visual          & Visual              & ALL                & Non-visual         & Visual             \\ \midrule
\multicolumn{1}{c|}{\multirow{4}{*}{\begin{tabular}[c]{@{}c@{}}Scene-driven\\ Knowledge\end{tabular}}}    & \multicolumn{1}{c|}{\multirow{4}{*}{ManipMob-MMKG}} & LOCKET-multimodal(ours)           & \textbf{\underline{84.9 (0.3)}} & \textbf{78.8 (0.5)} & 91.0 (0.2)          & \multirow{6}{*}{-} & \multirow{6}{*}{-} & \multirow{6}{*}{-} \\ \cmidrule(lr){3-6}
\multicolumn{1}{c|}{}                                                                                     & \multicolumn{1}{c|}{}                               & LOCKET-multimodal(no denoise)     & 83.9 (0.3)          & 77.6 (0.8)          & 90.2 (0.3)          &                    &                    &                    \\ \cmidrule(lr){3-6}
\multicolumn{1}{c|}{}                                                                                     & \multicolumn{1}{c|}{}                               & LOCKET-unimodal                   & 83.7 (0.1)          & 77.0 (0.4)          & 90.3 (0.4)          &                    &                    &                    \\ \cmidrule(lr){3-6}
\multicolumn{1}{c|}{}                                                                                     & \multicolumn{1}{c|}{}                               & LOCKET-multimodal (w/o QC\&R)     & 83.9 (0.2)          & 78.0 (0.7)          & 89.6 (0.7)          &                    &                    &                    \\ \cmidrule(r){1-6}
\multicolumn{1}{c|}{\multirow{2}{*}{\begin{tabular}[c]{@{}c@{}}Domain-specific\\ Knowledge\end{tabular}}} & \multicolumn{1}{c|}{LINK}                           & LINK~\cite{liu2023learning}                              & 83.1 (0.3)          & 76.0 (0.7)          & 90.1 (0.6)          &                    &                    &                    \\ \cmidrule(lr){2-6}
\multicolumn{1}{c|}{}                                                                                     & \multicolumn{1}{c|}{ManipMob-MMKG}                  & LOCKET-multimodal (scene-related) & 83.6 (0.1)          & 77.0 (0.4)          & 90.1 (0.3)          &                    &                    &                    \\ \midrule
\multicolumn{1}{c|}{\multirow{5}{*}{\begin{tabular}[c]{@{}c@{}}General \\ Knowledge\end{tabular}}}        & \multicolumn{1}{c|}{LegoFormer}                     & VLG~\cite{corona2022voxel}                               & \textbf{\underline{84.9 (0.3)}} & 78.4 (0.7)          & \textbf{91.2 (0.4)} & 79.0               & 71.7               & 86.0               \\ \cmidrule(l){2-9} 
\multicolumn{1}{c|}{}                                                                                     & \multicolumn{1}{c|}{ViLBERT}                        & ViLBERT~\cite{lu2019vilbert}                           & 83.1                & 76.6                & 89.5                & 76.6               & 73.0               & 80.2               \\ \cmidrule(l){2-9} 
\multicolumn{1}{c|}{}                                                                                     & \multicolumn{1}{c|}{BLIP-2}                         & BLIP-2~\cite{li2023blip}                            & 51.2                & 50.9                & 51.5                & -                  & -                  & -                  \\ \cmidrule(l){2-9} 
\multicolumn{1}{c|}{}                                                                                     & \multicolumn{1}{c|}{CLIP}                           & MATCH~\cite{thomason2022language}                             & 82.2 (0.4)          & 75.2 (0.7)          & 89.2 (0.9)          & 76.5 (0.5)         & 68.7 (0.9)         & 83.9 (0.5)         \\ \cmidrule(l){2-9} 
\multicolumn{1}{c|}{}                                                                                     & \multicolumn{1}{c|}{CLIP}                           & LAGOR~\cite{thomason2022language}                             & 82.6 (0.4)          & 75.3 (0.7)          & 89.8 (0.4)          & 77.0 (0.5)         & 69.4 (0.5)         & 84.3 (0.4)         \\ \bottomrule
\end{tabular}
}
\label{main_comparison_of_snare}
\end{table*}

For \textbf{RQ3}, to compare CKR-ManipMob-MMKG with CKR-ManipMob-MMKG (unimodal), we can find that the method using multimodal scene knowledge can achieve increasing performance obviously. Specifically, in seen scenes, CKR-ManipMob-MMKG has improvements of $1.35$ and $6.68$ in TL and SPL comparing to CKR-ManipMob-MMKG (unimodal) ($12.75 \rightarrow 11.40$ and $53.81 \rightarrow 60.49$). In unseen scenes, CKR-ManipMob-MMKG achieves $4.97$ improvement in TL ($25.35 \rightarrow 20.38$) and $2.39$ improvement in SPL ($13.74 \rightarrow 16.13$) respectively. Results show the advantages of multimodal scene knowledge in navigation planning compared to text-only scene knowledge.


Furthermore, within our analysis, we find that multimodal knowledge imports more useful information for embodied tasks while it also brings extra noise to interpret downstream modeling potentially. It is inevitable to further refine scene knowledge.
For \textbf{RQ4}, we propose \textit{Multimodal Denoising module} in Sec.~\ref{sec:gate}. We conduct an ablation study for multimodal denoising, and train a model named CKR-ManipMob-MMKG (no denoise) without using \textit{Multimodal Denoising module}.
Experimental results show the effectiveness of \textit{Multimodal Denoising module}. Specifically, in seen scenes, performance improves from $12.03$ and $55.24$ to $11.40$ and $60.49$ in TL and SPL respectively between CKR-ManipMob-MMKG (no denoise) and CKR-ManipMob-MMKG in TABLE~\ref{main_comparison}. In unseen scenes, the proposed CKR-ManipMob-MMKG model achieves improvements of $2.11$ and $3.63$ in TL and SPL, respectively, with the performance values $22.49 \rightarrow 20.38$ and $12.50 \rightarrow 16.13$. These results indicate the necessaries to refine multimodal scene knowledge for embodied tasks.


\subsection{3D Object Language Grounding Results Analysis}
\subsubsection{w/ knowledge vs. w/o knowledge}
For \textbf{RQ1}, we compare our proposed method with existing non-knowledge-enhanced methods,  shown in TABLE~\ref{main_comparison_of_snare}. Our proposed LOCKET-multimodal method achieves the best performance compared to MATCH and LAGOR. 
MATCH uses CLIP to encode the language reference and the image observation. The image feature embedding is mean-pooled and concatenated to the language embedding~\cite{thomason2022language}. 
LAGOR is based on the MATCH model and additionally regularized through the auxiliary task of predicting the canonical viewing angle of given multiple images.
MATCH and LAGOR are the 3D object language grounding models without external knowledge enhancement. 
In non-visual setting, our LOCKET-multimodal improves the accuracy by $3.1$ ($75.3 \rightarrow 78.4$), while in visual setting, achieves $1.1$ improvement ($89.8 \rightarrow 90.9$). 
The results indicate that our proposed LOCKET-multimodal can improve the embodied task performance obviously using external scene knowledge from ManipMob-MMKG. To be specific, the improvement in non-visual setting is more representative, in which the languages mainly describe the shapes or parts of the target object, such as ``\textit{big wooden doors}" and ``\textit{has two drawers}", instead of the name of the object in visual setting. Our ManipMob-MMKG complements insufficient or out-of-domain multimodal data for modeling. 

\begin{figure*}[tb]
\centering
\vspace{-0.3cm}  
\includegraphics[width=0.66\linewidth]{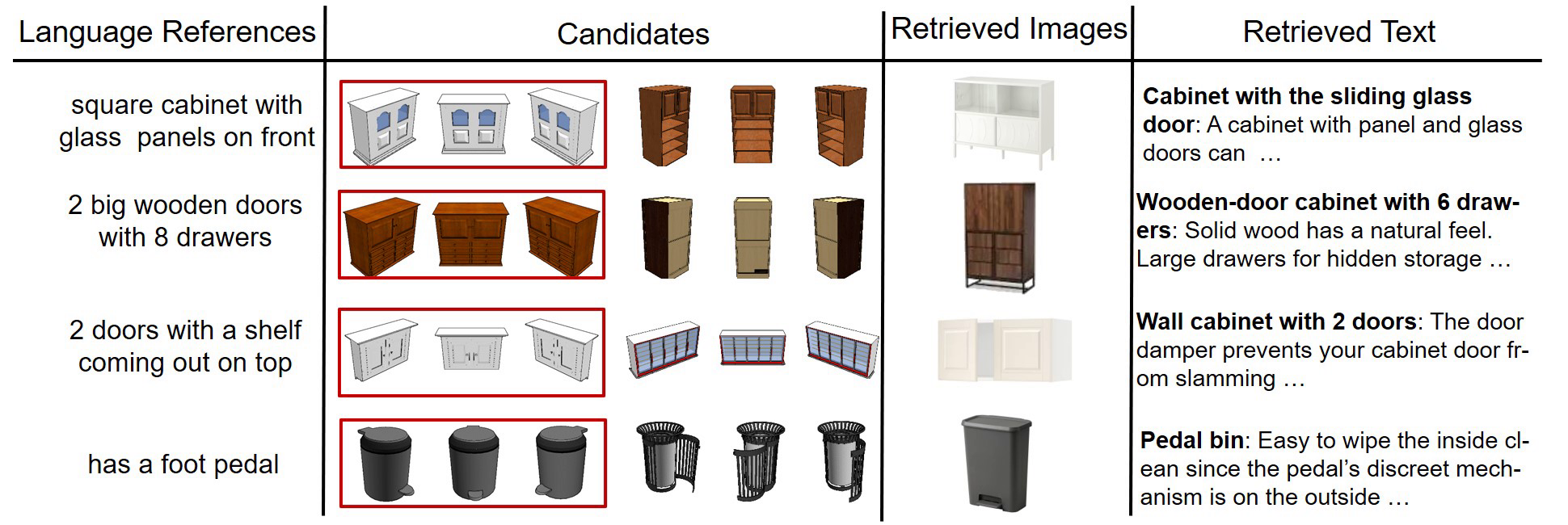}
\caption{ Visualization of 3D object language grounding cases and the process of scene knowledge retrieving.}
\label{fig:case_study_snare}
\end{figure*}

The usage process of our ManipMob-MMKG is shown in Fig.~\ref{fig:case_study_snare}. We provide four cases (lines) in non-visual setting, in which MATCH without using external knowledge achieves wrong prediction while our LOCKET-multimodal using ManipMob-MMKG predicts correctly.
The first column is the language references describing the part or the shape of the target object in the non-visual set. 
The second column is three views of two candidate objects including front-ahead, front-left, and front-right views. Objects in the red frame are ground-truth objects.
The last two columns are images and text knowledge retrieved from ManipMob-MMKG based on language references. 
Each visualization case (line) contains two groups of objects with three observations and one language reference about shape (square), material (wooden, glass), parts (shelf, drawers, doors, panels, foot pedal), etc. 
The agent can not choose the right object (in MATCH model) without external knowledge because the concepts in language references may not appear in the training set and non-visual concepts cannot generalize well in feature space. 
The retrieved images and text are equipped with the characteristics of the references and can be beneficial to the agent to predict the correct object.



\subsubsection{General or Domain-specific vs. Scene-driven}
\syx{
For \textbf{RQ2}, to compare models using general and domain-specific knowledge  in TABLE~\ref{tb:main_comparison_of_kb} with our scene knowledge used model LOCKET-multimodal, results shown in TABLE~\ref{main_comparison_of_snare} indicate our superiority in the embodied task. 
For methods using general knowledge in TABLE~\ref{main_comparison_of_snare} adopt pre-trained models (i.e. CLIP, ViLBERT~\cite{lu2019vilbert}, LegoFormer~\cite{yagubbayli2021legoformer}, BLIP-2~\cite{li2023blip}), of which are trained on the general, public dataset and provide implicit general parameter knowledge.
ViLBERT method~\cite{lu202012} fine-tunes 12-in-1 pre-trained ViLBERT~\cite{lu2019vilbert} as the backbone for MATCH instead of CLIP-ViT. 
Compared to CLIP-based method~\cite{thomason2022language}, ViLBERT extracts features from image regions and provides the ground-truth bounding boxes for each region.
VLG leverages explicit 3D prior knowledge from predicted volumetric voxel maps to improve language grounding performance by pre-trained 3D model LegoFormer~\cite{yagubbayli2021legoformer}. For domain-specific knowledge based methods, task-related multimodal symbolic knowledge is encoded and injected into the models. Our proposed LOCKET-multimodal outperforms $1.8$ ($84.9$ vs. $83.1$), $0.4$ ($78.8$ vs. $78.4$), and $39.5$ ($91.0$ vs. $51.5$) to conventional ViLBERT-based method~\cite{lu2019vilbert}, VLG~\cite{yagubbayli2021legoformer}, and BLIP-2~\cite{li2023blip} based method. Furthermore, to compare with domain-specific methods, as shown in TABLE~\ref{main_comparison_of_snare}, our proposed LOCKET-multimodal achieves over $1.0$ improvements in all metrics. In conclusion, all results show the superiority of our proposed scene knowledge injected method can achieve competitive performance with conventional general or domain-specific knowledge used methods in 3D object language grounding task. 
}

\subsubsection{Unimodal vs. Multimodal}
For \textbf{RQ3}, we compare the effectiveness of multimodal data and results are shown in TABLE~\ref{main_comparison_of_snare}. 
LOCKET-multimodal and LOCKET-unimodal represent LOCKET using image+text scene knowledge and text-only scene knowledge from ManipMob-MMKG respectively.
Compared LOCKET-unimodal with LOCKET-multimodal, ($83.6 \rightarrow 84.7$), we can find that multimodal scene knowledge promotes higher performance in 3D object language grounding task.
For \textbf{RQ4}, we conduct an ablation study to train a model removing \textit{Multimodal Denoising module} in LOCKET-multimodal named LOCKET-multimodal (no denoise).
To compare LOCKET-multimodal with LOCKET-multimodal (no denoise), the results show that accuracy achieves further improvement after refining the retrieved images from ManipMob-MMKG in all settings ($83.8 \rightarrow 84.7$, $77.1 \rightarrow 78.4$, $90.3 \rightarrow 90.9$), which indicate the necessaries to refine multimodal information in knowledge base construction and usage for embodied tasks again.


\subsection{Ablation Study of Scene-MMKG Construction}
\begin{figure}[tb]
\centering
\includegraphics[width=0.99\linewidth]{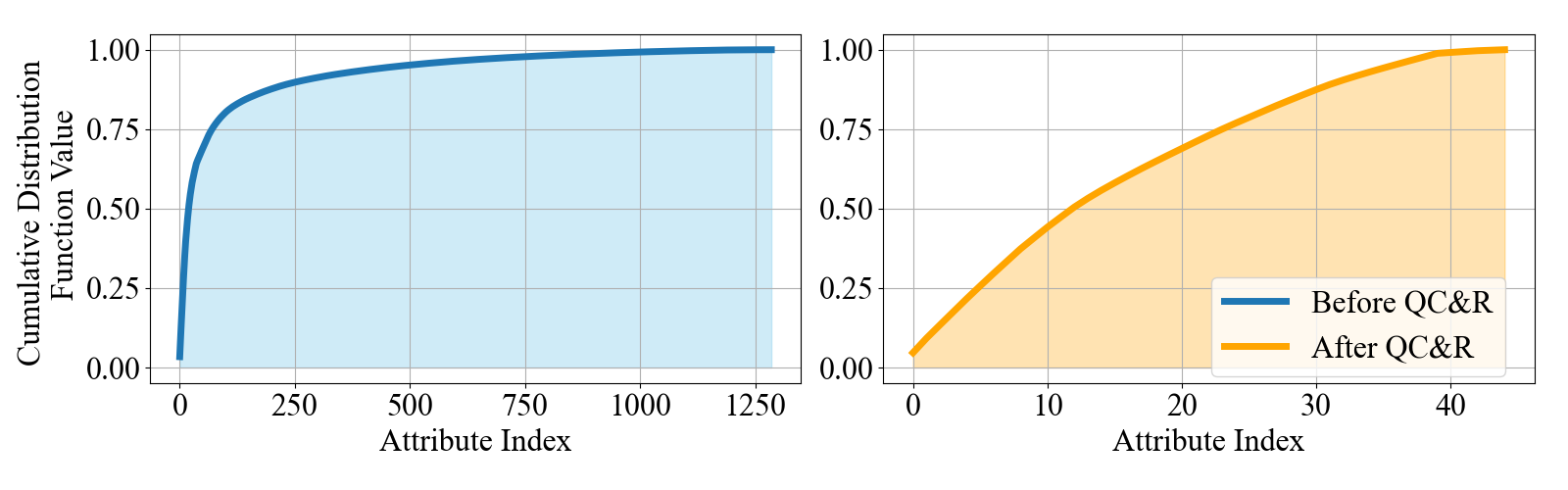}
\caption{\syx{Visualization of relieving long-tail problems by quality control and refinement.}}
\label{fig:long_tail}
\end{figure}

\syx{To further evaluate the performance of our proposed Scene-MMKG, we conduct an ablation study to analyze the effectiveness of QC\&R to relieve long-tail problems. The characteristic of ManipMob-MMKG without processing by QC\&R is shown in TABLE~\ref{tb:main_comparison_of_kb}. From Fig.~\ref{fig:long_tail}, the curves of the cumulative distribution function show that our proposed QC\&R could obviously reduce long-tail attributes with a uniform growth rate.
The embodied tasks performances of models using ManipMob-MMKG without processing by QC\&R are reported in TABLE~\ref{main_comparison} and TABLE~\ref{main_comparison_of_snare}, which quantitatively indicate that our proposed QC\&R promote the quality of scene knowledge and improve the knowledge-enhanced models.}

\section{Related Work}
Knowledge from the environment (world) forms the basis for how people make entailment and reasoning decisions~\cite {hofstadter2013surfaces}. In analogy with human, knowledge is vital for empowering the intelligence of a humanoid robot~\cite{liu2023survey}, which is a typical embodied agent. 
From the perspective of functionalities, The knowledge for embodied agent can be classified into two categories. The first is used to compensate for the insufficiency in the perception, such as ConceptNet~\cite{liu2004conceptnet}, WordNet~\cite{miller1995wordnet}, Wikipedia~\cite{vrandevcic2014wikidata}. The second is to enhance the decision-making basis and reasoning abilities, such as RoboBrain~\cite{saxena2014robobrain},
GPT series works~\cite{brown2020language,ChatGPTO40:online,openai2023gpt4}. We review these from two aspects: \textit{Symbolic Knowledge} and \textit{Parameter Knowledge}.

\subsection{Symbolic Knowledge}
Symbolic knowledge is usually explicit and constructed following the pre-defined manual schemas, which needs a scale of human efforts to collect and clean data but has advantages in explainability and human intervention.
It represents in the form of triples, tables, graphs, trees, and datasets directly, referring to perceptual knowledge, apperceptive knowledge, and expert system for embodied tasks.

\textbf{Perceptual Knowledge} contains shapes, sizes, appearance, weight, mass, density, materials, friction, etc., which directly affect the interactive modeling processing. 
ObjectFolder~\cite{gao2022objectfolder,gao2022objectfolder2} introduces a large-scale multisensory object dataset including vision, audio, touch with texture, material, and shape information for objects. 
WebChild~\cite{tandon2014webchild} and ConceptNet~\cite{liu2004conceptnet} contain commonsense triples that connect entities with adjectives via fine-grained relations.
3D scene graphs~\cite{rosinol20203d,armeni20193d,kim20193} are proposed to describe spatial perceptual knowledge in 3D scenarios to enhance embodied reasoning and planning tasks. 

\textbf{Apperceptive Knowledge} includes concepts, facts, commonsense, affordance, emotion, intention, event, etc., which cannot be perceived by agents directly and needs human experience and summary to abstract. COMET-ATMOIC-2020~\cite{hwang2021comet}, MLN-KB~\cite{zhu2014reasoning},
ADE-Affordance~\cite{chuang2018learning}, Lang-SHAPE~\cite{song2023learning} construct knowledge bases to reason about objects and their affordance for agent cognition and decision. Event evolutionary knowledge~\cite{hwang2021comet}, ATOMIC~\cite{sap2019atomic},
NEEG~\cite{li2018constructing} covers social and eventive aspects of inferential information, which can facilitate interpretable scene understanding and decision in real world. 

\textbf{Expert System} is used to guide the agent to imitate the decision-making ability of human experts in specific fields. RoboEarth~\cite{waibel2011roboearth}, Robobrain~\cite{saxena2014robobrain} and Open-EASE~\cite{tenorth2015open} design a knowledge engine, which learns and shares knowledge representations for robots for a variety of embodied tasks.

\subsection{Parameter Knowledge}
Knowledge representation can also use parameters of pre-trained models implicitly. Pre-trained models (PTM) predicate labels or generate tokes to provide implicit knowledge based on the superiority of training data, but they are also doubted in explainability. Representative pre-trained models used in embodied tasks include the language model, vision model, multimodal model, and robotic model.

\textbf{Language PTM} can be classified into traditional pre-trained language models and foundation models according to the scale of parameters. For traditional pre-trained language models, they use huge textual corpora to train discriminative models or generative models, such as BERT~\cite{devlin2019bert}, RoBERTa~\cite{liu2019roberta}, T5~\cite{raffel2020exploring}, BART~\cite{lewis2020bart}, GPT series works\cite{radford2019language,brown2020language,ChatGPTO40:online,openai2023gpt4} in natural language processing topic.
Foundation models~\cite{ChatGPTO40:online,openai2023gpt4,touvron2023llama,alpaca,HuggingChat,touvron2023llama, touvron2023llama2} (e.g. ChatGPT, Llama-2) are used to realize decision-making or complex scene understanding. 

\textbf{Vision PTM} provide visual perceptual knowledge by learnable parameters from pre-trained models for object detection~\cite{redmon2016you}, skeleton detection~\cite{liu2017skeleton}, human pose detection~\cite{wei2016convolutional}, scene graph generation~\cite{xu2017scene}, etc. For example, ImageNet~\cite{deng2009imagenet} or Visual Genome (VG)~\cite{krishna2017visual} are used to extract object semantics to enhance object-level scene understanding.


\textbf{Cross-modal PTM} is to make multimodal semantic alignment, such as vision-language models CLIP~\cite{radford2021learning}, ViLBERT~\cite{lu2019vilbert}, LXMERT~\cite{tan2019lxmert}. Typically, they are pre-trained on Flickr30k~\cite{plummer2015flickr30k}, ReferItGame~\cite{kazemzadeh2014referitgame}, RefCOCO ~\cite{yu2016modeling}, etc., and encode different modal inputs to the shared vector space and generate shared latent semantic representations~\cite{khandelwal2022simple,liang2022visual}. With foundation model rapid development, recently, generative multimodal models such as DALLE~\cite{ramesh2021zero}, DALLE2~\cite{ramesh2022hierarchical}, Point-E~\cite{nichol2022point}, which can generate visual knowledge based on abstract text descriptions and improve agent's reasoning abilities~\cite{li2023blip,ye2023mplugowl,zhu2023minigpt,liu2023llava} in embodied tasks. MiniGPT-4~\cite{zhu2023minigpt} has advanced multimodal generation capabilities like detailed image description and website creation. 

\textbf{Robot PTM} is investigated to pre-train a model for a robot specially and directly instead of using a pre-trained model from NLP or computer vision fields. There are two types of modeling lines. The first line is to train from scratch~\cite{kumar2022pre,nairr3m,majumdar2023we} using Internet data and task-specific data to train models for general robotic tasks (i.e. manipulation and mobility tasks), which achieve competitive zero-shot performance indicating the advantage of implicit knowledge in learnable parameters. The second line is to design a robotic model by fine-tuning or combining an existing model. PaLM-E~\cite{driess2023palm} co-trains an embodied large model based on 540B PaLM~\cite{chowdhery2022palm} and 22B ViT~\cite{dehghani2023scaling} and shows impressive performance in symbolic skill planning. ChatGPT~\cite{vemprala2023chatgpt} is adopted to design a user-on-the-loop robotic framework to realize decoupled decision-making.

\section{Conclusion}
In this paper, we investigate scene knowledge in the real world for embodied AI to help an agent understand situ environment and make correct decisions.
Specifically, we propose a novel scene-driven multimodal knowledge graph construction method to construct a scene-specific external knowledge base efficiently. A unified knowledge-enhanced framework for embodied tasks is designed. 
To evaluate the advantage of our proposed knowledge graph construction method for embodied tasks, we instantiate a multimodal knowledge graph considering robotic basic indoor abilities  (i.e. Manipulation and Mobility), named \textbf{ManipMob-MMKG}. 
Experiments and comparisons on characteristics of instantiated knowledge graph and downstream embodied tasks are performed. Comparisons indicate the broad advantages of instantiated knowledge base using our proposed construction method. Experimental results of downstream tasks show our ManipMob used methods outperform over $8$ (SPL) and $2$ (accuracy) than baselines without using knowledge respectively.

\section{Acknowledgments}
\noindent This work is supported by the National Natural Science Foundation of China (No.U21A20488, 62072323), Zhejiang Lab Open Research Project (No.K2022NB0AB04), Science and Technology Commission of Shanghai Municipality Grant (No. 22511105902) and Postdoctoral Fellowship Program of CPSF
(GZC20232292).



\vspace{-10 mm}
\begin{IEEEbiography}[{\includegraphics[width=1in,height=1.25in,clip,keepaspectratio]{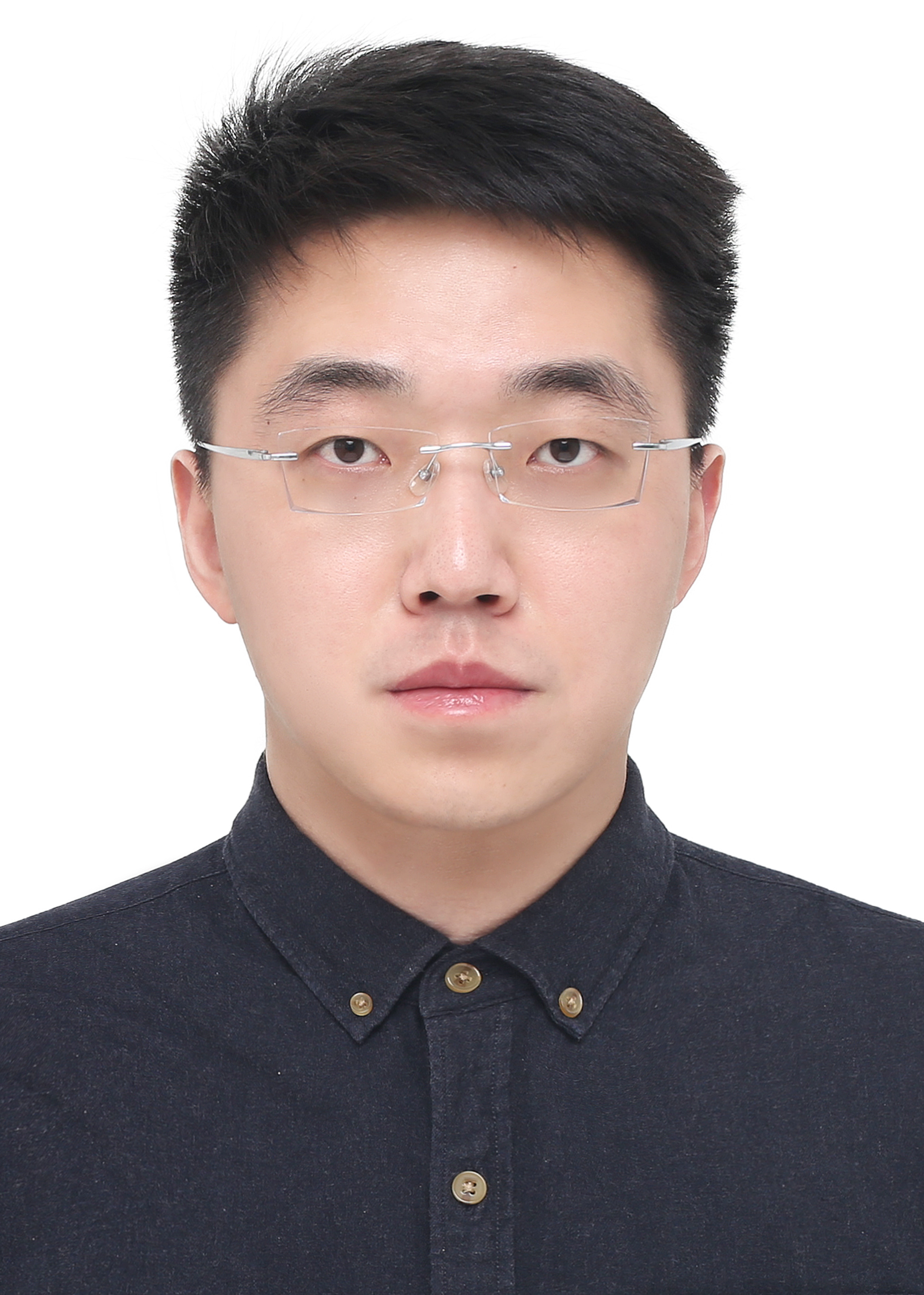}}]{Yaoxian Song} received his Ph.D. degree in computer science from Fudan University, Shanghai, China, jointly with Westlake University in 2023. His research interests include robotic grasping, embodied AI, multimodal knowledge graphs, visual language grounding, and unstructured scene understanding. He is currently a postdoctoral researcher at Zhejiang University, researching underwater scene cognition and decision-making.
\end{IEEEbiography}
\vspace{-11 mm}

\begin{IEEEbiography}[{\includegraphics[width=1in,height=1.25in,clip,keepaspectratio]{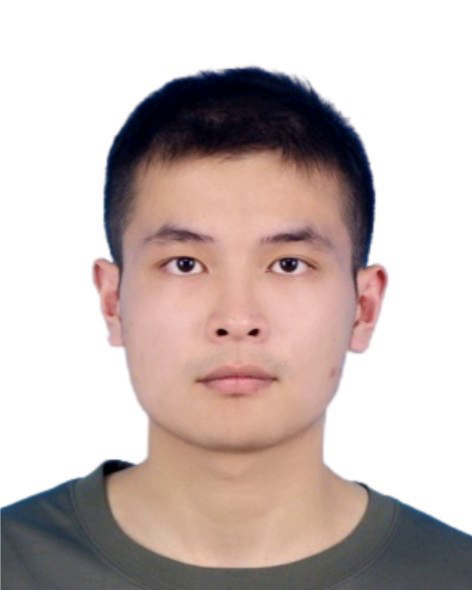}}]{Penglei Sun} received his MSc degree in Computer Technology from Fudan University, Shanghai, China, in 2023. He is currently a Ph.D. student in Hong Kong University of Science and Technology. His research interests include knowledge graph and multimodal learning.
\end{IEEEbiography}
\vspace{-11 mm}

\begin{IEEEbiography}[{\includegraphics[width=1in,height=1.25in,clip,keepaspectratio]{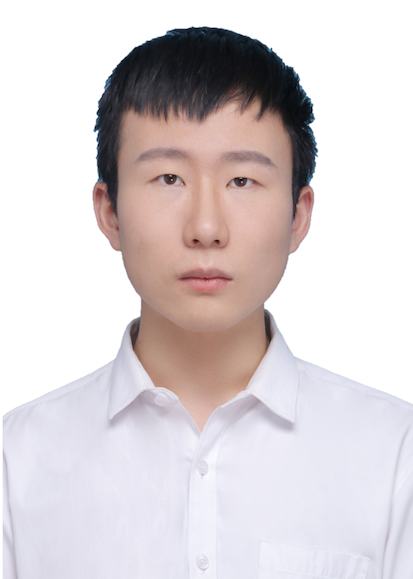}}]{Haoyu Liu}
received the BSc and MSc degrees in computer science from Chang 'an University and Xi'an Jiaotong-Liverpool University, respectively. He is currently with the Zhejiang Laboratory. His research interests include knowledge graphs, cross-modal retrieval, and embodied AI.
\end{IEEEbiography}
\vspace{-11 mm}

\begin{IEEEbiography}[{\includegraphics[width=1in,height=1.25in,clip,keepaspectratio]{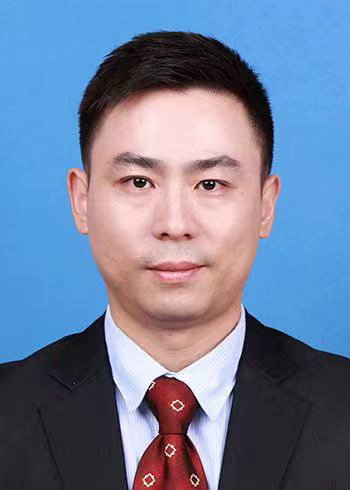}}]{Zhixu Li}
is a professor with the School of Computer Science at Fudan University, China. He used to be a professor at Soochow University between 2014 and 2021. He received his Ph.D. degree in Computer Science from the University of Queensland in 2013. His main research interests are Data \& Knowledge Engineering, and Cognitive Intelligence, and he is particularly interested in Multi-modal Knowledge Graph and Cross-Modal Cognitive Intelligence
 \end{IEEEbiography}
\vspace{-11 mm}

\begin{IEEEbiography}[{\includegraphics[width=1in,height=1.25in,clip,keepaspectratio]{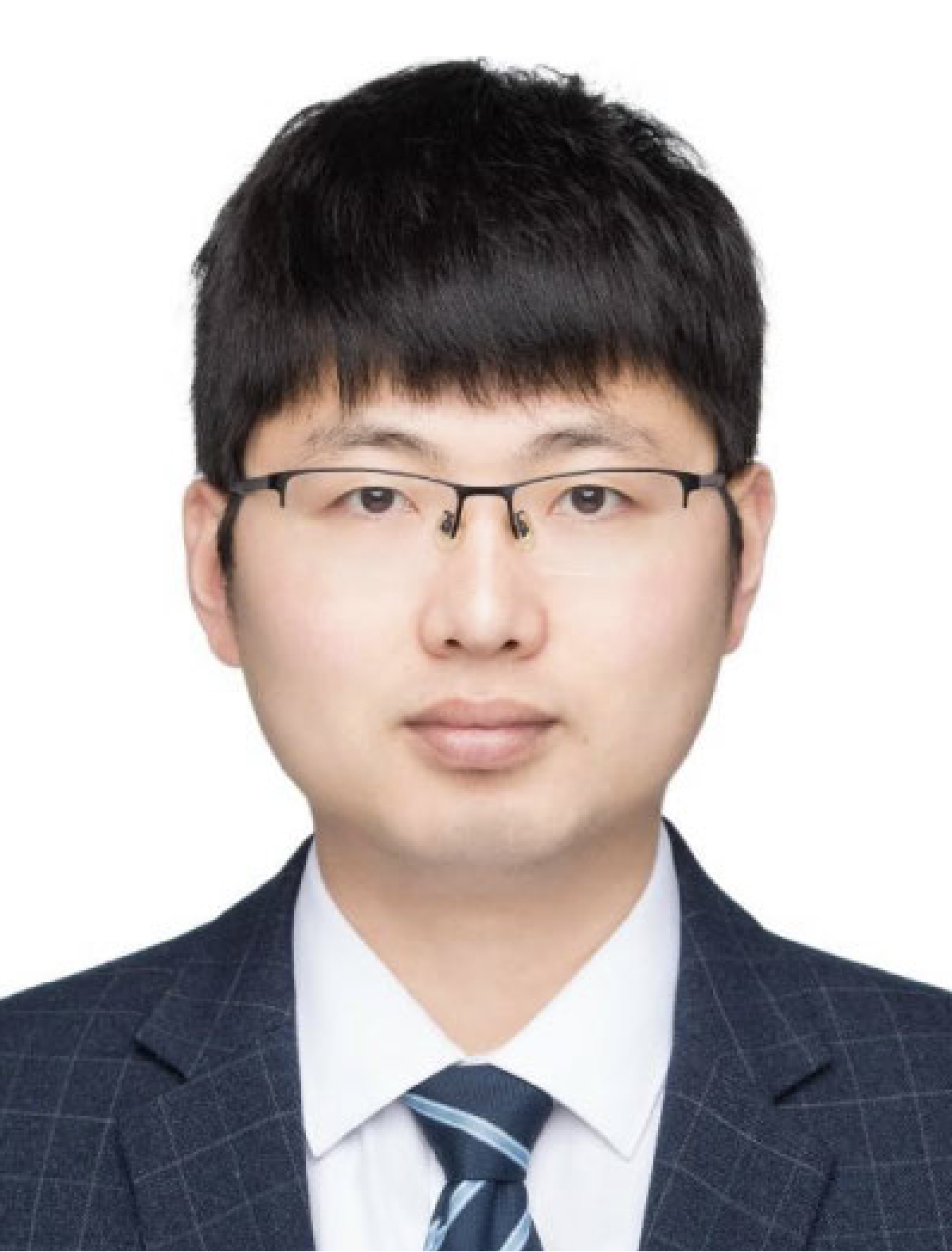}}]{Wei Song}received the B.Eng. and Ph.D. degrees in mechatronic control engineering from Zhejiang University, Zhejiang, China, in 2008 and 2013, respectively. He is currently with the Zhejiang Laboratory. His research interests include marine bionic soft robot, underwater manipulator driven by ropes, wall climbing robot, and deep water hydraulic equipment.
\end{IEEEbiography}
\vspace{-11 mm}

\begin{IEEEbiography}[{\includegraphics[width=1in,height=1.25in,clip,keepaspectratio]{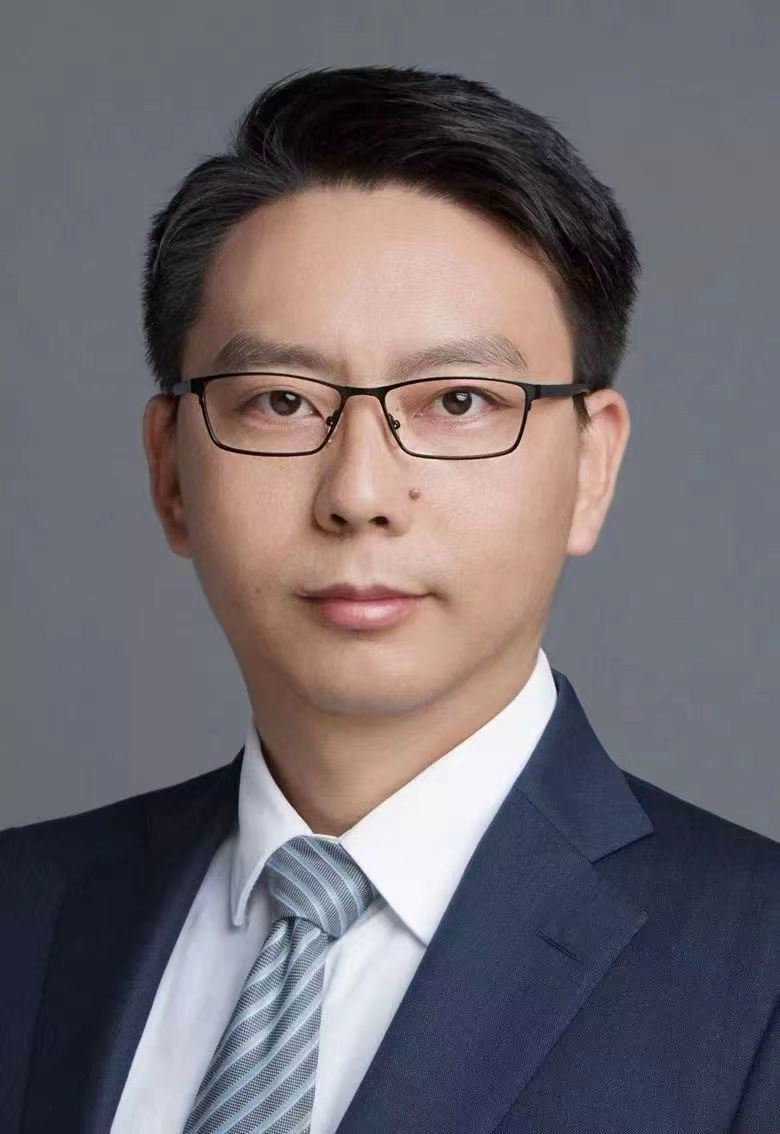}}]{Yanghua Xiao}is a professor of computer science at Fudan University. He is the director of Knowledge Works Lab, Fudan University. He got his Ph.D. degree in software theory from Fudan University, Shanghai, China, in 2009. His research interests include big data management and mining, graph database, knowledge graph. He was a visiting professor of Human Genome Sequencing Center at Baylor College Medicine, and visiting researcher of Microsoft Research Asia.
\end{IEEEbiography}
\vspace{-10 mm}

\begin{IEEEbiography}[{\includegraphics[width=1in,height=1.25in,clip,keepaspectratio]{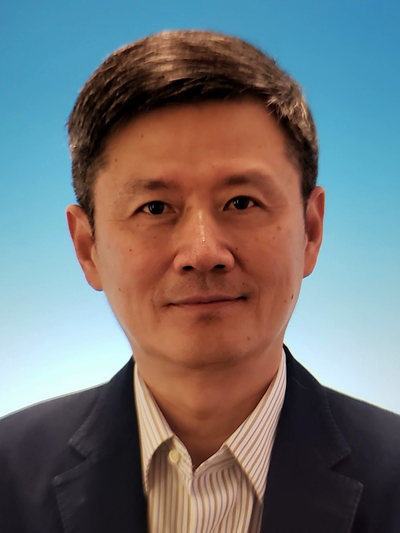}}]{Xiaofang Zhou}
(Fellow, IEEE) received the BSc and MSc degrees from Nanjing University, in 1984 and 1987, respectively, and the PhD degree in computer science from the University of Queensland in 1994. He is currently a chair professor with the Department of Computer Science and Engineering, Hong Kong University of Science and Technology. His research interests include web information systems, data mining, and data quality management.
\end{IEEEbiography}

\end{document}